\pdfoutput=1
\documentclass{article}

\PassOptionsToPackage{dvipsnames}{xcolor}

\usepackage{microtype}
\usepackage{graphicx}
\usepackage{subcaption}
\usepackage{booktabs}
\usepackage{fancyvrb}
\usepackage{url}
\usepackage{hyperref}
\usepackage[preprint]{icml2026}
\hypersetup{
     colorlinks = true,
     linkcolor = BrickRed,
     anchorcolor = blue,
     citecolor = Blue,
     filecolor = blue,
     urlcolor = Green
}
\makeatletter
\renewcommand{\Notice@String}{}
\makeatother
\makeatletter
\g@addto@macro{\UrlBreaks}{\do\a\do\b\do\c\do\d\do\e\do\f\do\g\do\h\do\i\do\j\do\k\do\l\do\m\do\n\do\o\do\p\do\q\do\r\do\s\do\t\do\u\do\v\do\w\do\x\do\y\do\z\do\A\do\B\do\C\do\D\do\E\do\F\do\G\do\H\do\I\do\J\do\K\do\L\do\M\do\N\do\O\do\P\do\Q\do\R\do\S\do\T\do\U\do\V\do\W\do\X\do\Y\do\Z\do\0\do\1\do\2\do\3\do\4\do\5\do\6\do\7\do\8\do\9}
\makeatother
\usepackage{amsmath}
\usepackage{amssymb}
\usepackage{mathtools}
\usepackage{xcolor}
\usepackage{tikz}
\usetikzlibrary{positioning,fit,arrows.meta,calc,backgrounds}
\definecolor{jblue}{HTML}{0072B2}
\definecolor{jorange}{HTML}{E69F00}

\newcommand{\pmci}[2]{#1\,\textcolor{gray}{$\pm$\,#2}}
\newcommand{\ciint}[3]{#1\,\textcolor{gray}{[#2,\,#3]}}
\usepackage{xspace}
\newcommand{\jb}{OpenDecisions\xspace}
\newcommand{\pub}{JevBench-Open}
\newcommand{\jeffL}{Jeff-1.0-Large}
\newcommand{\cmark}{\checkmark}

\usepackage{fontawesome5}

\icmltitlerunning{Releasing the Inner Jev: A Simple Recipe for Open Decision Models}

\begin{document}

\twocolumn[
  \icmltitle{Releasing the Inner Jev: A Simple Recipe for Open Decision Models}

  \icmlsetsymbol{equal}{*}
  \begin{icmlauthorlist}
    \icmlauthor{Jonathan Geuter}{equal,harvard,kempner}
    \icmlauthor{Andrew Cheng}{equal,harvard}
    \icmlauthor{David Alvarez-Melis}{harvard,kempner,msr}
  \end{icmlauthorlist}

  \icmlaffiliation{harvard}{Harvard University}
  \icmlaffiliation{kempner}{Kempner Institute}
  \icmlaffiliation{msr}{Microsoft Research}
  \icmlcorrespondingauthor{Jonathan Geuter}{jonathan.geuter@gmx.de}

  \icmlkeywords{decision models, calibration, LoRA, Jev}
  \vskip 0.3in
]

\printAffiliationsAndNotice{\icmlEqualContribution}

\begin{abstract}
LLM-based decision-making systems, also called System One models, have recently gained popularity, most notably through Jev~\citep{almeida2026jev}. Given a typed input state, a question, and a set of options, they return a calibrated probability for each option in a single forward pass, without text generation. We show that open instruction-tuned LLMs can achieve performance comparable to Jev \textit{without any additional training}, by reading out answers from letters (A, B, C, etc.) in the model's output logits at the last token position. Next, we propose a simple supervised LoRA fine-tuning recipe that can improve their performance further.
We evaluate four instruction-tuned LLMs: Gemma~4~31B, Gemma~4~12B, Qwen3.6-27B, and Qwen3.5-4B. Furthermore, we introduce \textit{\jb}, a novel 2,352-question validation dataset designed to reflect the task composition of JevBench~\citep{jevbench} and which computes a Capability score based on accuracy and calibration. Without fine-tuning, Gemma 4 31B scores 79.6 on \jb, compared to Jev's 80.1. After fine-tuning, it scores 89.0, outperforming Jev by 8.9 Capability points and on par with the strongest open decision model we evaluated. The same recipe improves the three other models by 7.8 to 15.7 points. On a suite of public decision datasets, our models achieve accuracy and calibration comparable to Jev. Our results show that open LLMs already possess System One capabilities, and with simple fine-tuning they outperform Jev on a wide range of decision tasks. We release our models, training and serving code, \jb, and \textit{DecisionBank}, our training dataset consisting of 261,561 decisions.

\end{abstract}

\vspace{-3em}
\begin{center}
\small
\href{https://huggingface.co/collections/jgeuter/jeff-11-6ada6708f3ed49beaf0f97ae}{\faCubes~Jeff 1.1 models}\quad
\href{https://github.com/j-geuter/Jeff}{\faGithub~Jeff}\\
\href{https://huggingface.co/datasets/jgeuter/OpenDecisions}{\faDatabase~OpenDecisions}\quad
\href{https://huggingface.co/datasets/jgeuter/DecisionBank/viewer/DecisionBank-260k}{\faDatabase~DecisionBank}
\end{center}

\begin{figure*}[t]
  \centering
  \resizebox{\textwidth}{!}{\begin{tikzpicture}[
  font=\sffamily\small,
  card/.style={draw=black!25, rounded corners=5pt, fill=black!2, inner sep=6pt},
  chip/.style={draw=black!30, rounded corners=6pt, fill=white, inner xsep=4pt, inner ysep=2pt, font=\sffamily\footnotesize},
  letter/.style={circle, anchor=center, fill=black!70, text=white, inner sep=0pt, minimum size=3.6mm, font=\sffamily\scriptsize\bfseries},
  frozen/.style={draw=black!35, fill=black!10, rounded corners=1.2pt, inner sep=0pt},
  trained/.style={draw=jblue, fill=jblue!25, rounded corners=1.2pt, inner sep=0pt},
  arr/.style={-{Stealth[length=2.4mm]}, line width=1pt, black!45},
  lbl/.style={font=\sffamily\footnotesize, text=black!60, align=center},
]
\node[anchor=north west, text width=3.55cm, align=left, execute at begin node={\hyphenpenalty=10000\relax}, font=\sffamily\footnotesize, text=black!65] (state) at (0,0)
  {Order \#4411 was charged twice; the customer wants the money back.};
\node[anchor=north west, font=\sffamily\footnotesize\bfseries] (question) at ([yshift=-1.2mm]state.south west)
  {Which team should handle it?};
\node[chip, anchor=north west] (o1) at ([yshift=-1.6mm]question.south west) {\tikz[baseline=-0.75ex]\node[letter]{A}; billing};
\node[chip, right=1mm of o1] (o2) {\tikz[baseline=-0.75ex]\node[letter]{B}; tech};
\node[chip, right=1mm of o2] (o3) {\tikz[baseline=-0.75ex]\node[letter]{C}; refund};
\begin{scope}[on background layer]\node[card, fit=(state)(question)(o1)(o3)] (req) {};\end{scope}
\node[lbl, below=1.5mm of req] (reqlbl) {state, question, options};

\coordinate (m) at ($(req.east)+(1.05,0)$);
\foreach \i in {0,...,4} {
  \node[frozen, minimum width=15mm, minimum height=2.8mm, anchor=west] (L\i) at ($(m)+(0,{0.78-0.39*\i})$) {};
  \node[trained, minimum width=2.8mm, minimum height=2.8mm, anchor=west] (A\i) at ($(L\i.east)+(0.7mm,0)$) {};
}
\node[font=\sffamily\scriptsize, text=black!55, anchor=south] (lt) at ([yshift=0.6mm]L0.north) {layers};
\node[font=\sffamily\scriptsize\bfseries, text=jblue, anchor=south] (at) at ([yshift=0.6mm, xshift=1.2mm]A0.north) {LoRA};
\begin{scope}[on background layer]
  \node[draw=black!25, rounded corners=4pt, fill=white, inner sep=2.2mm, fit=(L0)(A0)(L4)(A4)(lt)(at)] (llm) {};
\end{scope}
\node[font=\sffamily\bfseries, anchor=south] at ([yshift=0.4mm]llm.north) {LLM};
\node[lbl, below=1.5mm of llm] {one forward pass};
\draw[arr] (req.east) -- (req.east -| llm.west);
\coordinate (llmE) at (llm.east |- req.east);
\coordinate (hc) at ($(llmE)+(0.95,0)$);
\foreach \i/\c in {0/40,1/22,2/62,3/14,4/46,5/24} {
  \node[draw=black!30, fill=jorange!\c, minimum width=3mm, minimum height=3mm, inner sep=0pt] (h\i) at ($(hc)+(0,{0.75-0.3*\i})$) {};
}
\node[anchor=south] at (h0.north) {$h$};
\node[lbl, anchor=north] at ([yshift=-1mm]h5.south) {last hidden\\state};
\draw[arr] (llmE) -- ($(hc)+(-2.6mm,0)$);
\coordinate (dh) at ($(hc)+(1.25,0)$);
\foreach \i/\l in {0/A,1/B,2/C} {
  \node[letter] (dl\i) at ($(dh)+(0,{0.42-0.42*\i})$) {\l};
  \node[frozen, minimum width=11mm, minimum height=3.2mm, anchor=west] (dr\i) at ($(dl\i.east)+(1mm,0)$) {};
  \node[anchor=west, inner sep=0.6pt] (dp\i) at (dr\i.east) {$+$};
  \node[trained, minimum width=3.2mm, minimum height=3.2mm, anchor=west] (dd\i) at (dp\i.east) {};
}
\draw[arr] ($(hc)+(2.6mm,0)$) -- ($(dl1.west)+(-1mm,0)$);
\coordinate (sb) at ($(dd1.east)+(1.15,-0.62)$);
\draw[black!40] ($(sb)+(-0.08,0)$) -- ++(1.92,0);
\foreach \i/\v/\f/\l in {0/0.81/jblue/A,1/0.07/jblue!40/B,2/0.12/jblue!40/C} {
  \fill[\f] ($(sb)+({0.68*\i},0)$) rectangle ++(0.4,{1.45*\v});
  \node[font=\sffamily\scriptsize, anchor=south, text=black!70] at ($(sb)+({0.68*\i+0.2},{1.45*\v})$) {\v};
  \node[font=\sffamily\scriptsize\bfseries, anchor=north] at ($(sb)+({0.68*\i+0.2},-0.04)$) {\l};
}
\draw[arr] ($(dd1.east)+(1.2mm,0)$) -- ($(sb)+(-0.2,0.62)$);
\node[lbl, anchor=north] at ($(sb)+(0.88,-0.42)$) {softmax\\over options};
\node[draw=jblue, fill=jblue!8, rounded corners=5pt, align=center, inner sep=4pt] (ans) at ($(sb)+(3.4,0.62)$)
  {\textbf{billing}\\[-1pt]{\footnotesize $p = 0.81$}};
\draw[arr] ($(sb)+(1.98,0.62)$) -- (ans.west);
\node[lbl, below=1.5mm of ans] {answer};
\node[font=\sffamily\scriptsize, text=black!55, anchor=south] at ([yshift=1mm]dr0.north) {LM-head rows};
\node[font=\sffamily\scriptsize, text=jblue, anchor=south] at ([yshift=1mm]dd0.north) {$\Delta$};
\node[lbl, anchor=north] at ([yshift=-2mm]$(dr2.south)!0.5!(dd2.south)$) {decision head};
\node[frozen, minimum width=3mm, minimum height=3mm, anchor=north west] (lg1) at ([yshift=-3.5mm]reqlbl.south -| req.west) {};
\node[lbl, anchor=west] (lg1t) at ([xshift=1mm]lg1.east) {frozen};
\node[trained, minimum width=3mm, minimum height=3mm, anchor=west] (lg2) at ([xshift=5mm]lg1t.east) {};
\node[lbl, anchor=west] at ([xshift=1mm]lg2.east) {trained: LoRA adapters and head correction $\Delta$};
\end{tikzpicture}}
  \caption{\textbf{A Jev-style decision in one forward pass.} The request is rendered as a multiple-choice prompt with
  lettered options. The LLM reads it once; the decision head maps the last hidden state $h$ to one logit per option,
  starting from the language-model head's rows for the letters, so the untrained model is the zero-shot readout.
  Training adds LoRA adapters and a learned correction $\Delta$ to the head. A softmax over the real options gives the probabilities.}
  \label{fig:overview}
\end{figure*}

\section{Introduction}\label{sec:intro}
 Many applications require choosing among a fixed set of options, such as routing a support ticket, checking a request against a policy, or picking the right tool in an agentic workflow. Language models can make these decisions by generating text, but generating tokens sequentially introduces additional inference costs and latency. For tasks requiring only a choice and its associated probabilities, a single forward pass avoids this overhead and provides uncertainty estimates that downstream systems can use directly.

Using a pretrained language model with a classification head has long been well-established in the literature, including ULMFiT~\citep{howard2018ulmfit}, GPT~\citep{radford2018gpt}, and BERT~\citep{devlin2019bert}. Jev~\citep{almeida2026jev} adopts this approach: given a state, a question, and a set of options, it returns a probability for each option. TypeSafe, the company behind Jev, describes it as using a specialized architecture, a parallel sampler, and a training method called Reinforcement Learning for Calibrated Decisions (RLCD)~\citep{almeida2026jev}. Its weights and training recipe are not public. The format has attracted several open implementations, which are compared on the JevBench leaderboard~\citep{jevbench}.

\textit{Are specialized architectures or training methods necessary for competitive decision models?} We find that a simple approach using open-source language models already achieves performance comparable to Jev. When given a multiple-choice question, we extract probabilities over the options from the model's logits (see Section~\ref{sec:method} for details), an established approach to evaluating multiple-choice models~\citep{hendrycks2021mmlu,robinson2023mcq}, and apply simple temperature scaling for calibration. Without further fine-tuning, Gemma~4~31B~\citep{gemma4} and Qwen3.6-27B~\citep{qwen36} achieve Capability scores of 79.6 and 79.0 on our validation set \textit{\jb} respectively, compared with 80.1 for Jev. Both models also achieve similar average accuracy to Jev across seven public decision datasets.

Next, we investigate whether supervised fine-tuning can improve performance further. We add LoRA adapters to the language model and initialize a decision head from the open-source model's slice of the language model head corresponding to the answer letter tokens (A, B, C, etc.). Both are trained jointly using cross-entropy on DecisionBank, our 261,561 decision question training dataset that we built from public datasets as well as synthetically generated samples. Our recipe also includes model souping~\citep{wortsman2022soups} across three training seeds, and bidirectional attention for most models. With our training recipe, Gemma~4~31B improves from 79.6 to 89.0 Capability (Jeff-1.1-Large). This exceeds Jev by \pmci{8.9}{2.3} points and matches the strongest open decision model we evaluated, StartLux-27B Instant~\citep{startlux} (88.5). These results show that existing language models can match and exceed Jev on decision tasks using a simple supervised training procedure, without a specialized architecture or reinforcement learning.

Our contributions are:
\begin{itemize}
\item \textbf{Existing language models are competitive decision models.} We show that without fine-tuning, open instruction-tuned language models achieve comparable performance to Jev. This holds on both \jb and seven publicly available decision datasets (Section~\ref{sec:zeroshot}).

\item \textbf{A validation set for decision models.}
We introduce \jb, a validation dataset for decision models containing 2,352 questions across 11 task families, three question types, four difficulty levels, and 23 languages. Its labels are checked programmatically or by a panel of three models, and its scores broadly reproduce the rankings of systems on the official JevBench leaderboard (Section~\ref{sec:data}).

\item \textbf{A simple training recipe and training data for decision models.} We introduce DecisionBank, 261,561 decisions with soft targets, and a simple supervised LoRA training procedure. It improves Gemma~4~31B by \pmci{9.4}{2.0} Capability points on \jb, exceeding Jev and matching the strongest open decision model, and improves all four models we train. Ablations show that LoRA drives most of the improvement, while training the decision head alone provides little to no benefit (Sections~\ref{sec:method}, \ref{sec:training} and~\ref{sec:ablations}).

\end{itemize}

\section{Background and related work}\label{sec:background}

\paragraph{Jev-style decision models.}
A Jev-style decision consists of a state (text or JSON), a question, and $K$ possible options. JevBench considers three types of questions: \emph{choice} (pick one of $K$ options), \emph{yes/no} (do the given criteria hold for the
state?), and \emph{score} (pick one of several ordered levels). For
yes/no questions,  the prediction is represented by a single probability $P(\text{yes})$. These probabilities allow downstream systems not only to select an answer, but also to return a confidence associated with it. For this to be reliable, the predicted probabilities should be well-calibrated.

\paragraph{Classification with pretrained LLMs.}
LLM classification predates Jev: ULMFiT adds a classification head to a pretrained recurrent language model, while GPT and BERT adapt pretrained transformers using supervised output layers~\citep{howard2018ulmfit,radford2018gpt,devlin2019bert}. PET~\citep{schick2021pet} uses cloze-style prompts to construct soft labels and then trains a conventional sequence classifier with a classification head. A separate line of work handles labels supplied in natural language: \citet{yin2019zeroshot} formulate zero-shot classification as entailment between the input and candidate label hypotheses, while GLiClass~\citep{stepanov2025gliclass} jointly encodes the input and candidate labels and scores them in a single pass.

\paragraph{Open decision models.} Several open systems implement this format using existing language models. Some require no additional training and obtain probabilities directly from answer-letter logits, including~SemIf~\citep{semif}, open-alternative-jev~\citep{openaltjev}, Cygnet~\citep{cygnet} and
deck31b~\citep{deck31b}; others average over option orders~\citep{anyjev} or score each option in its own
prompt~\citep{llm2jev}. Fine-tuned systems include Plumb~\citep{plumb}, JevK5~\citep{jevk5}, Hopper~\citep{hopper} and
reflex~\citep{reflex} (LoRA on Qwen3.5-4B), decider~\citep{decider} (full fine-tuning on about 95 datasets),
Winnow-12B~\citep{winnow}, Torchcast-decision-12B~\citep{torchcast}, Quyet-1.0~\citep{quyet} (LoRA on Gemma~4~31B),
and several fine-tunes of Qwen3.8-27B: Eikos~\citep{eikos}, AutoJev~\citep{autojev}, Clef~\citep{clef} and
Open-Jev~\citep{openjev}; newer systems include StartLux-Decision~\citep{startlux}, pplx-decider~\citep{pplxdecider}, Chon~\citep{chon} and Wald~\citep{wald}. Despite the variety of approaches, few of these systems report evaluations beyond the JevBench leaderboard.

\paragraph{Concurrent work.} \citet{li2026llmasjev} also study whether existing language models can perform Jev-style decisions without additional training. Rather than using answer letters, they extract probabilities from numbered options using a partially completed answer prompt. They find that  an unmodified Qwen3.5-4B performs comparably to other decision models built on the same backbone, which
corroborates our zero-shot finding. They also study fine-tuning 4B and 0.6B models with a listwise loss with KL regularization. We compare their number readout with our letter readout on \jb\ and find similar zero-shot Capability scores for Gemma~4~31B (80.5 for numbered options and 79.6 for answer letters). Since the number readout requires multiple forward passes (each digit requires a forward pass, plus one to denote the end of the number), whereas our letter readout requires only one, we adopt the latter.

\paragraph{Calibration.}
Correct predictions alone are not enough when downstream systems rely on the reported probabilities. Neural networks are often overconfident, and temperature scaling provides a simple way to improve calibration without changing the preferred answer~\citep{guo2017calibration,desai2020calibration}. Previous work shows that pretrained language models can produce reasonably calibrated probabilities, while instruction tuning can increase overconfidence~\citep{kadavath2022know,tian2023justask}. Predictions can also depend on the order and labels of the answer options~\citep{zheng2024selectors,pezeshkpour2023order}. In this work, we show that the model output probabilities themselves already lead to better calibration than Jev (see Section \ref{sec:results}). We leave more sophisticated calibration approaches for future work.

\section{\jb: a validation set for Jev-style models}\label{sec:data}

\paragraph{Why a new dataset.}
Evaluating decision models requires (i) enough questions to reliably compare their performance and (ii) sufficient coverage of the tasks they are designed to solve. Existing public datasets have limitations in both respects. JevBench has released only 231 questions, which we call \pub. Other public decision datasets, such as those from SemIf and TypeSafe, contain only 100 to 350 questions as well. Although larger multiple-choice benchmarks exist, they primarily test general knowledge rather than decisions based on supplied information. To address these limitations, we introduce \jb, a 2,352-question validation set designed to reflect the task composition of JevBench.

\paragraph{Coverage.} \jb\ has 2,352 decisions in 11 families (Table~\ref{tab:families}): routing requests to models or
teams, multilingual decisions, fairness pairs, business use cases, counting and arithmetic traps, negation pairs,
option-order and perturbation pairs, composite policy rules, and questions with exact answer probabilities.
It has all three question types (1,178 choice, 586 yes/no and 588 score questions), four difficulty tiers (easy,
standard, judge, hard) and 23 languages (78\% English). Its mix of types, tiers, families and languages follows the
published composition of the public JevBench~v1.6 split.

\begin{table}[t]
  \caption{\textbf{\jb\ families and label sources.} Programmatic items are written and labeled by code; real requests are user requests from public datasets. Calibration: questions with exact answer probabilities.}
  \label{tab:families}
  \centering\footnotesize
  \setlength{\tabcolsep}{3pt}
  \begin{tabular}{@{}lrrrr@{}}
    \toprule
    Family & Code & LLM-written & Real & Total \\
    \midrule
    LLM routing & 22 & 391 & 281 & 694 \\
    Multilingual & & 393 & 29 & 422 \\
    Fairness pairs & & 236 & & 236 \\
    Business use cases & 20 & 193 & & 213 \\
    Counting traps & 162 & & & 162 \\
    Negation pairs & 160 & & & 160 \\
    Option order & & 120 & & 120 \\
    Perturbation pairs & & 110 & & 110 \\
    Composite rules & 106 & & & 106 \\
    Calibration, generated & 94 & & & 94 \\
    Calibration & 35 & & & 35 \\
    \midrule
    Total & 599 & 1,443 & 310 & 2,352 \\
    \bottomrule
  \end{tabular}
\end{table}

\paragraph{Labels.} The items come from three sources. 599 are written and labeled by code. 1,443 were written by LLM authoring agents (Claude Opus 4.8 and 5.5; \citealp{claudeopus}) from fixed briefs and then confirmed by a
panel of three open models (Qwen3.8-27B, gpt-oss-120b and GLM-4.7-Flash; \citealp{qwen38,gptoss,glm47}) that answer each
item blind, with reasoning. An item was kept only if all three agreed with the author's answer, or two of three did and their average probability on that answer was at least 0.5. 310 are real user requests (Chatbot Arena~\citep{chiang2024arena},
Dolly~\citep{conover2023dolly}, GSM8K~\citep{cobbe2021gsm8k}, MATH-500~\citep{hendrycks2021math,lightman2023verify}), again answered by the three-model panel, and kept only when all models agree.

\paragraph{Quality checks.}
Some questions are constructed in pairs, such as the same question with its answer options reordered or with a detail of the input changed. We remove both questions if either fails verification by the three-model panel. We also compare candidate questions against our training data and the publicly released JevBench questions to check for overlap, removing 43 samples.

\paragraph{Validity.} 14 public decision models
come out on \jb\ in nearly the order of JevBench~\citep{jevbench} (Spearman correlation of 0.90). Four of these systems were never used
while building \jb; a linear fit on the other ten predicts their leaderboard scores with an error of 3.0 points. \pub\
and a validation set drawn from our training distribution predict the leaderboard much worse (Appendix~\ref{app:jb16}).

\paragraph{Scoring.} We score a system by $\text{Capability} = (I + C)/2$, adapted from JevBench's scorer without its
abstention band for yes/no answers. Intelligence $I$ is chance-corrected accuracy, $100\,(\text{acc} - \text{chance})/(1
- \text{chance})$, computed per question type and difficulty tier and averaged with tier weights 0.1, 0.2, 0.3 and 0.4
(easy to hard) and equal type weights. A yes/no answer counts as correct when its more likely side is correct.
Calibration $C$ averages one score per type: $100\,(1 - \text{ECE}/0.5)$~\citep{naeini2015ece} of the top answer on
hard choice items, of $P(\text{yes})$ on yes/no items, and for score items the mean of the top-answer ECE score and
$100\,(1 - \text{RPS})$, the ranked probability score~\citep{epstein1969rps}. We report 95\% intervals as \pmci{value}{1.96\,SE}. The standard
error SE combines question sampling (a paired bootstrap over question groups, 1,000 resamples, the same for all
systems) and, for trained models, the variation between training seeds.

\section{A simple recipe}\label{sec:method}
In this section, we explain how decision models work in detail, and give a simple training recipe to apply on pretrained LLMs to improve their decision-making capabilities.

\paragraph{Letter readout.} Each sample is presented to the model with options labeled A, B, C, and so on. Following SemIf~\citep{semif},  we use a JSON message containing the state (\texttt{evidence}),
question (\texttt{criterion}), and answer options. The model is instructed to select one answer letter, using its standard chat template with thinking disabled (see Appendix~\ref{sec:requests} for details).

Rather than generating a textual answer, we extract the model's logits
for the answer letters after a single forward pass. Let
$h\in\mathbb{R}^d$ denote the final hidden state at the last
prompt position, after the final normalization layer, and let
$W_0\in\mathbb{R}^{16\times d}$ contain the rows of the
language-model head corresponding to the letters A through P.
For a question with $K$ options, we obtain the probability
of option $k$ by applying a softmax over these $K$ logits:
\begin{equation}\label{eq:readout}
  p_k =
  \frac{\exp\big((W_0 h)_k\big)}
  {\sum_{j=1}^{K}\exp\big((W_0 h)_j\big)},
  \qquad k=1,\ldots,K.
\end{equation}
This converts the scores for the valid answer letters into a
probability distribution over the available options. Each letter
is represented by a single token, so all probabilities can be
obtained in one forward pass. For Gemma, we retain its original
final-logit soft-capping before applying softmax, as described
below. Yes/no questions are represented by two options, while
scoring questions use ordered answer levels. The readout supports
up to 16 options.

\paragraph{Decision head.}
Rather than learning a new decision head from scratch, we exploit the pretrained language model's answer-letter readout and add a trainable correction:
\begin{equation}\label{eq:head}
  z = W_0\,h + A\,\mathrm{std}(h) + b,
\end{equation}
followed by the softmax over the $K$ options. Here, $W_0$ are the fixed letter rows from Eq.~\eqref{eq:readout},
$\mathrm{std}(h)$ standardizes $h$ with mean and standard deviation computed once on training data before training (which are kept fixed during training), and the learned correction
$A \in \mathbb{R}^{16 \times d}$, $b \in \mathbb{R}^{16}$ starts at zero. For Gemma, we preserve the model's original final-logit soft-capping by replacing $W_0 h$ with $c \tanh(W_0 h / c)$ with $c = 30$. Without soft-capping,
the decision head is a single affine function of $h$. After training, the LoRA adapter can be merged into the pretrained weights, so inference requires only one forward
pass.

\paragraph{Calibration.}
The probabilities produced by language models are not always well calibrated~\citep{guo2017calibration,desai2020calibration,tian2023justask}. In our experiments, Gemma's probabilities are particularly overconfident. We address this using temperature scaling~\citep{guo2017calibration}, which adjusts the predicted probabilities without changing the preferred answer. For our released models, we fit one temperature for each of the three question types by minimizing negative log-likelihood on held-out task families that are not in the training data. We keep temperatures at least one, so calibration can reduce confidence but not increase it. In addition, our serving code moves a yes/no answer with $0.2 < P(\text{yes}) < 0.8$ to the nearer of 0.199 and 0.801. We note that this is a very simple calibration strategy, and more sophisticated calibration approaches, such as directly incorporating calibration in the training loss, could likely help.

\begin{table}[t]
  \caption{\textbf{DecisionBank} (261,561 questions), the training data of Jeff~1.1. Teacher: Qwen3.8-27B with reasoning, mean of two samples.}
  \label{tab:data}
  \centering\small
  \begin{tabular}{@{}p{3.15cm}rp{2.35cm}@{}}
    \toprule
    Part & Rows & Target \\
    \midrule
    decider (79 public tasks, custom questions) & 163,732 & label and teacher (50/50) \\
    Human votes (4 datasets) & 9,168 & votes and teacher (50/50) \\
    Written by LLMs & 60,219 & teacher \\
    Programmatic & 28,442 & exact (soft or one-hot) \\
    \midrule
    Total & 261,561 & \\
    \bottomrule
  \end{tabular}
\end{table}

\paragraph{Training data.} We train on DecisionBank (Table~\ref{tab:data}), 261,561 questions that combine public datasets, questions written by a language model, and programmatic questions. The largest part, 163,732 questions, comes from decider's public training mixture (79 tasks, at most 2,094 questions per task) and its custom questions~\citep{decider}. We also add four public datasets with human vote distribution labels: Civil Comments~\citep{borkan2019civil}, Measuring Hate Speech~\citep{kennedy2020hate}, STS-B~\citep{cer2017sts}, and ChaosNLI~\citep{nie2020chaosnli}. We also found it beneficial to mix the labels with soft labels from a teacher model. Therefore, we generate soft labels with Qwen3.8-27B~\citep{qwen38} as the mean of two answers with reasoning. For all public data, the target is the average of the dataset's label (or vote distribution) and the teacher. Furthermore, we synthetically generated 60,219 questions with language models (Qwen3.8-27B, and Qwen3.6-27B for a small part), where a model generates a sample with an intended answer without reasoning, and we only keep samples where the model would generate the intended answer when using reasoning. The remaining 28,442 questions come from the code that generates \jb's programmatic questions, run with new random seeds. No training question has the state of a \jb\ question, and we also filtered for n-gram overlap (Appendix~\ref{app:train-eval}).

\paragraph{Training.}
We fine-tune the models using supervised learning, keeping the original language-model weights frozen. Specifically, we add rank-32 LoRA adapters ($\alpha=64$, no dropout) to all attention and MLP projections, including the linear-attention projections in the Qwen models, and train them jointly with the decision head. We minimize cross-entropy against the target probability distribution and also shuffle the order of the answer options during training.

We use AdamW~\citep{loshchilov2019adamw} with learning rates of $2\cdot10^{-5}$ for LoRA and $10^{-4}$ for the decision head, and a WSD learning rate scheduler over 1500 steps, where each step processes 65,536 padded tokens. Except for Gemma~4~31B, we use bidirectional attention, since the decision is read at the last prompt position and no text is generated. We train three runs per model with different seeds and use the model soup across all three. We use no reinforcement learning, reward models, or auxiliary losses. Further details are provided in Appendix~\ref{app:train}, including details about small adaptations to the training recipe we found helpful for the 4B model.

\paragraph{Reasoning.} With Jeff~1.1, we also support reasoning. Optionally, if the calibrated confidence (i.e., the probability of the most likely option) of the one-pass answer is below a threshold for its question type, the base model (the backbone without the LoRA adapter) writes a short reasoning of at most 512 tokens (256 for Qwen3.5-4B). The decision model (with LoRA adapter) then reads the original prompt, concatenated with the reasoning and ``Final answer:'', and its prediction on this updated prompt is averaged with the distribution of the first pass (mean of the log-probabilities) and calibrated with its own temperature per question type. The reasoning path is efficiently implemented with vLLM for the base model weights in practice. We note that a fixed threshold on the model's confidence is a primitive rule, that is suboptimal particularly for samples where the true distribution is meant to be soft; we leave more sophisticated reasoning pipelines for future research.

\paragraph{Changes from Jeff 1.0.} Our first suite of models (Jeff~1.0) used the same architecture and loss, but a different training pool with teacher labels from Qwen3.5-122B-A10B and Qwen3.6-27B on a subset of questions, no programmatic questions, only 400 training steps, no model souping, causal attention for all models, no support for reasoning, and different calibration temperatures. Appendix~\ref{app:jeff10} compares the two versions in more detail; the ablations in Section~\ref{sec:ablations} still use the Jeff~1.0 recipe.

\section{Results}\label{sec:results}

\paragraph{Setup.}
We evaluate four instruction-tuned open models: Gemma~4~31B and 12B~\citep{gemma4}, Qwen3.6-27B~\citep{qwen36}, and Qwen3.5-4B~\citep{qwen35}. We evaluate each without any additional training ("zero-shot"), as well as after applying our training recipe outlined in the previous section. We release the four trained models as Jeff-1.1-Large (Gemma~4~31B), Jeff-1.1-Qwen-27B, Jeff-1.1-Medium (Gemma~4~12B), and Jeff-1.1-Qwen-4B.

We compare them with Jev, accessed through its API, and 20 public systems using their authors' implementations. All systems are evaluated on the same questions in \jb. For each public system, we report the better of its original predictions and predictions recalibrated using our separate calibration set, based on its \jb\ score. This favors the competing systems. Since we selected our training recipe using \jb, we also evaluate on seven held-out decision datasets, \pub\ (a small 231-sample public dataset released by JevBench), and six standard multiple-choice benchmarks.
\begin{figure}[t]
  \centering
  \includegraphics[width=\columnwidth]{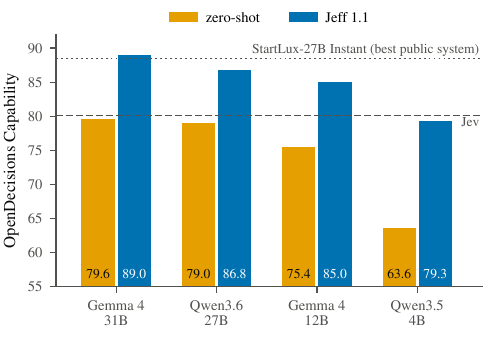}
  \caption{\textbf{LLMs are on par with Jev without any additional fine-tuning; after fine-tuning they are significantly stronger than Jev.} \jb\ Capability of the four open models
  without training (orange) and after training (Jeff-1.1). Dashed: Jev.
  Dotted: the strongest public system (StartLux-27B Instant). The two largest models reach Jev's level zero-shot,
  and training moves the three larger models clearly beyond it.}
  \label{fig:main}
\end{figure}

\subsection{LLMs as zero-shot decision models}
\label{sec:zeroshot}

We find that strong open language models already achieve performance comparable to Jev without fine-tuning. On \jb, Gemma~4~31B achieves a Capability score of 79.6 ($I=71.9$, $C=87.4$) and Qwen3.6-27B scores 79.0, compared with 80.1 for Jev (Figure~\ref{fig:main}, Table~\ref{tab:main}). Both models also achieve similar average accuracy to Jev across seven publicly available decision datasets (Section~\ref{sec:objective}).

Calibration plays an important role in this result. Gemma's raw answer-letter probabilities are overconfident, and temperature scaling alone raises its Capability score from 72.9 to 79.6. The fitted temperatures are 6.0 for choice questions and 7.2 for yes/no questions (Appendix~\ref{app:calib}). The answer-letter readout also discards very little probability mass: Gemma~4~31B assigns 99.99\% of its next-token probability to the available answer letters, and every backbone we evaluate assigns over 99.4\% on average (Appendix~\ref{app:readout}).

Performance also depends on the underlying model. The smaller Gemma~4~12B and Qwen3.5-4B models score 75.4 and 63.6 on \jb, respectively. On the public decision datasets, zero-shot accuracy increases rapidly with model size up to 4B parameters, and approaches Jev's level between 12B and 31B (Appendix~\ref{app:scaling}). The Gemma~4~31B base model evaluated using a plain completion prompt scores only 64.0, while the instruction-tuned model scores 78.9 with the same prompt (Appendix~\ref{app:readout}). Instruction tuning cannot be replaced by our recipe either: the base model reaches only 70.3 after training with our recipe (Section~\ref{sec:ablations}).

\begin{table*}[t]
  \caption{\textbf{Main results.} \jb\ Intelligence, Calibration and Capability, and Capability and top-1 accuracy on
  \pub\ (the 231 items JevBench released publicly). Zero-shot and Jeff-1.1: letter readout with temperatures fitted on held-out
  data. Public systems: as served or with fitted temperatures, whichever scores higher on
  \jb. Trained: \cmark\ fine-tuned, -- frozen readout, ? unknown. Bold: best per column. Public systems: subset of all system we evaluated;
  Table~\ref{tab:temps-public} lists all 21.}
  \label{tab:main}
  \centering\footnotesize
  \setlength{\tabcolsep}{5pt}
  \begin{tabular}{@{}llcccccc@{}}
    \toprule
    & & & \multicolumn{3}{c}{\jb} & \multicolumn{2}{c}{\pub} \\
    \cmidrule(lr){4-6}\cmidrule(lr){7-8}
    System & Backbone & Trained & $I$ & $C$ & \textbf{Capability} & Capability & Accuracy (\%) \\
    \midrule
    \multicolumn{8}{@{}l}{\emph{Ours}} \\
    Zero-shot & Gemma 4 31B & -- & 71.9 & 87.4 & 79.6 & 73.4 & 91.8 \\
    Jeff-1.1-Large & Gemma 4 31B & \cmark & 83.5 & \textbf{94.5} & \textbf{89.0} & 82.2 & 93.1 \\
    Zero-shot & Qwen3.6-27B & -- & 67.5 & 90.5 & 79.0 & 77.1 & 88.3 \\
    Jeff-1.1-Qwen-27B & Qwen3.6-27B & \cmark & 81.2 & 92.5 & 86.8 & \textbf{83.6} & 90.0 \\
    Zero-shot & Gemma 4 12B & -- & 64.0 & 86.8 & 75.4 & 72.0 & 86.1 \\
    Jeff-1.1-Medium & Gemma 4 12B & \cmark & 77.2 & 92.8 & 85.0 & 83.2 & 88.7 \\
    Zero-shot & Qwen3.5-4B & -- & 46.1 & 81.2 & 63.6 & 63.4 & 80.1 \\
    Jeff-1.1-Qwen-4B & Qwen3.5-4B & \cmark & 68.4 & 90.2 & 79.3 & 72.7 & 85.7 \\
    \midrule
    \multicolumn{8}{@{}l}{\emph{Public systems}} \\
    StartLux-27B Instant & Qwen3.8-27B & \cmark & \textbf{85.2} & 91.7 & 88.5 & 81.4 & \textbf{93.9} \\
    pplx-decider-v1.3 & Qwen3.8-27B & \cmark & 79.2 & 91.1 & 85.1 & 78.4 & 90.9 \\
    StartLux-4B Instant & Qwen3.5-4B & \cmark & 72.8 & 91.6 & 82.2 & 77.4 & 88.7 \\
    Quyet-1.0-Large & Gemma 4 31B & \cmark & 76.7 & 86.9 & 81.8 & 78.8 & 90.9 \\
    deck31b & Gemma 4 31B & -- & 73.6 & 89.0 & 81.3 & 80.5 & 93.1 \\
    Jev 1.13.0 (API) & closed & \cmark & 71.9 & 88.3 & 80.1 & 73.6 & 86.6 \\
    pplx-decider-flash-v1.3 & Qwen3.5-4B & \cmark & 66.4 & 81.9 & 74.1 & 74.0 & 84.4 \\
    Torchcast-decision-4B & Qwen3.5-4B & \cmark & 63.7 & 83.2 & 73.5 & 72.4 & 83.5 \\
    \bottomrule
  \end{tabular}
\end{table*}

\begin{table*}[t]
  \caption{\textbf{Performance on seven held-out public decision datasets.}
ECE: mean expected calibration error across the six datasets with hard labels (excluding TypeSafe); lower is better.}
  \label{tab:heldout}
  \centering\footnotesize
  \setlength{\tabcolsep}{5pt}
  \begin{tabular}{@{}lccccccccc@{}}
    \toprule
    System & TREC & CB & WANLI & Synth. & Pert. & TypeSafe & decider & \textbf{Average} & ECE~$\downarrow$ \\
    Items & 500 & 56 & 256 & 144 & 108 & 348 & 7,800 & 9,212 & \\
    \midrule
    Jev 1.13.0 (API) & 92.4 & 82.1 & 76.6 & 97.2 & \textbf{100.0} & 87.0 & 80.6 & 88.0 & .066 \\
    \midrule
    Gemma 4 31B, zero-shot & 92.8 & 85.7 & 78.9 & 95.1 & 97.2 & \textbf{87.5} & 81.9 & 88.4 & .077 \\
    Jeff-1.1-Large & 92.8 & 78.6 & \textbf{80.5} & 96.5 & \textbf{100.0} & 87.0 & 82.3 & 88.2 & .062 \\
    Qwen3.6-27B, zero-shot & 93.0 & 83.9 & 77.7 & \textbf{97.9} & \textbf{100.0} & 84.3 & 81.4 & 88.3 & \textbf{.052} \\
    Jeff-1.1-Qwen-27B & \textbf{93.6} & 87.5 & \textbf{80.5} & 97.2 & \textbf{100.0} & 84.6 & \textbf{82.6} & \textbf{89.4} & .064 \\
    Gemma 4 12B, zero-shot & 92.2 & \textbf{92.9} & 71.5 & 95.1 & 99.1 & 84.9 & 78.9 & 87.8 & .080 \\
    Jeff-1.1-Medium & 92.6 & 78.6 & 76.6 & 95.8 & \textbf{100.0} & 80.8 & 79.5 & 86.3 & .065 \\
    Qwen3.5-4B, zero-shot & 87.4 & 73.2 & 63.7 & 80.6 & 77.8 & 80.7 & 74.1 & 76.8 & .094 \\
    Jeff-1.1-Qwen-4B & 90.6 & 82.1 & 77.0 & 92.4 & 98.1 & 75.1 & 79.6 & 85.0 & .071 \\
    \bottomrule
  \end{tabular}
\end{table*}

\subsection{Training goes beyond Jev}\label{sec:training}

Our training raises Gemma~4~31B from 79.6 to 89.0 Capability on \jb\ (Jeff-1.1-Large), a gain of \pmci{9.4}{2.0}. This is the highest \jb\ score of all systems we evaluated (Table~\ref{tab:main}). Jeff-1.1-Large leads Jev by \pmci{8.9}{2.3}, and is slightly ahead of StartLux-27B Instant (\pmci{0.5}{1.6}), the best open source model on \jb we evaluated. It has the best Calibration of all systems (94.5); StartLux-27B Instant has slightly higher Intelligence (85.2 vs.\ 83.5).

The same recipe improves every model we trained: Qwen3.6-27B by \pmci{7.8}{2.1}, Gemma~4~12B by \pmci{9.6}{2.2}, and Qwen3.5-4B by \pmci{15.7}{2.9}. Compared with our first release, Jeff~1.1 is \pmci{4.1}{1.7} points higher for the 31B model and gains the most for the smaller models (Appendix~\ref{app:jeff10}). On the standard multiple-choice benchmarks, training raises the accuracy of every model as well (Section~\ref{sec:objective}).

\begin{figure*}[t]
  \centering
  \includegraphics[width=0.8\textwidth]{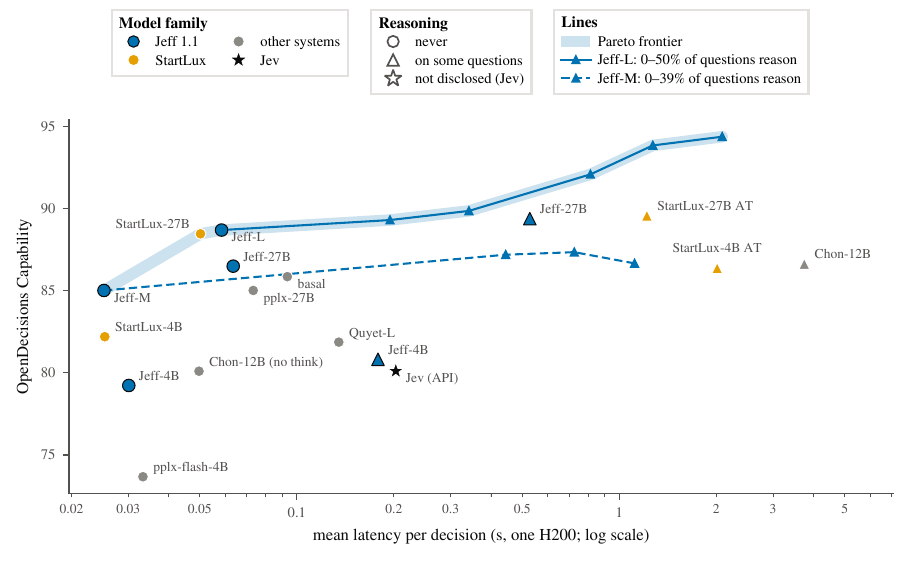}
  \caption{\textbf{\jb Capability against mean latency}. One H200 GPU per system. Lines: Jeff-1.1-Large and Jeff-1.1-Medium with a growing share of
  questions that reason.}
  \label{fig:lat-mean}
\end{figure*}

\paragraph{Reasoning when unsure.} With reasoning (Section~\ref{sec:method}), Jeff-1.1-Large scores 92.5 on \jb\ when 16.5\% of the questions reason, with 41 output tokens per question on average, and 94.1 when half of the questions reason (108 tokens per question), against 89.0 in one forward pass. Reasoning raises both Intelligence (83.5 to 89.4 and 93.3) and Calibration (94.5 to 95.7 and 94.9). Figure~\ref{fig:lat-mean} shows the Capability-latency pareto frontier defined by our 12B model, as well as the 31B with various degrees of reasoning.
However, lower reasoning thresholds that result in more overall reasoning can significantly increase mean latency, from 58ms per decision on Jeff-1.1-Large without reasoning to around two seconds per decision at the highest level of reasoning we tested, which amounts to 50\% of questions. Appendix~\ref{app:latency} describes further details, including median latencies, output tokens, and faster serving options for reasoning.

\subsection{Public evaluation sets}\label{sec:objective}

\jb\ was built by us and we partially used it for training hyperparameter selection. We also evaluate our models on a suite of public datasets.

\paragraph{Held-out decision datasets.} Table~\ref{tab:heldout} reports accuracy and calibration on seven public decision
datasets containing 9,212 questions that were not included in our training data. The datasets are TREC~\citep{li2002trec} (question classification), CB~\citep{demarneffe2019cb} and WANLI~\citep{liu2022wanli} (natural language inference), SemIf's synthetic and perturbation sets~\citep{semif} (Synth.\ and Pert.\ in Table~\ref{tab:heldout}), TypeSafe's public decision examples, scored by agreement with the most likely reference answer, and 29 held-out classification tasks of decider~\citep{decider}, averaged across tasks (Appendix~\ref{app:public}).
Averaged over the seven sets, Jeff-1.1-Large scores 88.2, Jeff-1.1-Qwen-27B 89.4 and Jev 88.0; every open model of about 30B parameters, trained or not, is at or above Jev. On the largest set, 7,800 items from the held-out tasks of decider, our models are clearly ahead of Jev: \pmci{1.7}{0.7} for Jeff-1.1-Large and \pmci{2.0}{0.7} for Jeff-1.1-Qwen-27B. Calibration is similar: the mean ECE over the six hard-label sets is 0.062 for Jeff-1.1-Large and 0.066 for Jev. Jev is best on SemIf's synthetic set (97.2, tied with Jeff-1.1-Qwen-27B), which is public and may have been used in its training.

\begin{table}[t]
  \caption{\textbf{What matters} (Jeff~1.0 recipe, Gemma 4 31B, \jb\ Capability; 400 steps and seed 1 unless noted). $\Delta$: difference to the
  recipe's three-seed mean (84.4), $\pm$ 1.96 standard errors (question sampling and seed variation). Pretrained head:
  trained first on frozen features.}
  \label{tab:ablation}
  \centering\footnotesize
  \setlength{\tabcolsep}{3pt}
  \begin{tabular}{@{}lcl@{}}
    \toprule
    Variant & Capability & $\Delta$ \\
    \midrule
    Recipe (mean of 3 seeds) & 84.4 & \\
    No training (zero-shot) & 79.6 & \pmci{$-$4.8}{2.0} \\
    \midrule
    \multicolumn{3}{@{}l}{\emph{What is trained}} \\
    Head only, frozen backbone & 78.9 & \pmci{$-$5.5}{2.5} \\
    LoRA, head frozen at its initialization & 84.9 & \pmci{$+$0.6}{2.1} \\
    \midrule
    \multicolumn{3}{@{}l}{\emph{Backbone}} \\
    Gemma base model + LoRA recipe & 70.3 & \pmci{$-$14.1}{3.3} \\
    \midrule
    \multicolumn{3}{@{}l}{\emph{Head initialization}} \\
    LoRA, pretrained head, trained jointly & 82.5 & \pmci{$-$1.9}{2.4} \\
    LoRA, pretrained head kept frozen & 81.6 & \pmci{$-$2.8}{2.4} \\
    \midrule
    \multicolumn{3}{@{}l}{\emph{LoRA rank}} \\
    Rank 16 & 84.7 & \pmci{$+$0.3}{2.1} \\
    Rank 64 & 84.5 & \pmci{$+$0.1}{2.2} \\
    Rank 128 & 84.9 & \pmci{$+$0.6}{2.2} \\
    \midrule
    \multicolumn{3}{@{}l}{\emph{Training data}} \\
    Public data only (no teacher labels) & 82.5 & \pmci{$-$1.9}{2.2} \\
    Hard labels instead of soft targets & 84.1 & \pmci{$-$0.2}{2.1} \\
    \bottomrule
  \end{tabular}
  \vspace{-1em}
\end{table}

\paragraph{\pub.} On the 231 items JevBench released publicly (Table~\ref{tab:main}), Jeff-1.1-Large answers 93.1\% correctly and scores 82.2 Capability, against 86.6\% and 73.6 for Jev. Without training, Gemma~4~31B already answers 91.8\% correctly. Among public systems, StartLux-27B Instant scores 93.9\% / 81.4 Capability and deck31b 93.1\% / 80.5. The concurrent work
of \citet{li2026llmasjev} reports 81.4\% for its training-free and 84.0\% for its LoRA-trained 4B model. With 231 items,
one system's Capability has a bootstrap standard deviation of about 3 points, so this set does not faithfully separate strong models.

\paragraph{JevBench leaderboard.} On the official JevBench leaderboard~\citep{jevbench} (v1.8.10, November 2026), Jeff-1.1-Medium (81.9 Capability) is the best open model with at most 12B parameters, and Jeff-1.1-Qwen-4B (77.3) is tied for the best open model with at most 4B parameters with Torchcast Decision 4B (77.5). Jeff-1.1-Large scores 85.9, which places it fourth among the ranked open models.\footnote{The leaderboard has evaluated Jeff-1.1-Large, but not yet officially ranked it. Our JevBench submissions include reasoning on a small number of samples. Run the same way on \jb, they score 90.3 (Jeff-1.1-Large), 87.1 (Jeff-1.1-Medium) and 81.2 (Jeff-1.1-Qwen-4B), against 89.0, 85.0 and 79.3 in one forward pass.} Jev scores 77.1 (Appendix~\ref{app:public}, Table~\ref{tab:board}), so even our 4B matches Jev, while our larger models outperform it. Each system is measured on its own draw of items.

\paragraph{Knowledge-heavy multiple choice.} On six standard benchmarks (BoolQ, MMLU, MMLU-Pro, ARC-Challenge,
WinoGrande, SciQ; \citealp{clark2019boolq,hendrycks2021mmlu,wang2024mmlupro,clark2018arc,sakaguchi2020winogrande,welbl2017sciq})
Jev is ahead: 92.9 average accuracy against 89.5 for Jeff-1.1-Large, a gap of \pmci{3.3}{0.8}, mostly on
MMLU-Pro, MMLU and WinoGrande (Appendix~\ref{app:public}). These tasks test general knowledge more than reading and reasoning about a given state. Jeff-1.1-Large and Jeff-1.1-Qwen-27B (89.5 and 89.6) lead the strongest public systems (Quyet-1.0-Large 88.3, deck31b 88.1), and training raises accuracy for all four of our models and lowers the 31B model's calibration error by 3x (ECE 0.110 to 0.035).

\subsection{Ablations}\label{sec:ablations}

\begin{figure}[t]
  \centering
  \includegraphics[width=\columnwidth]{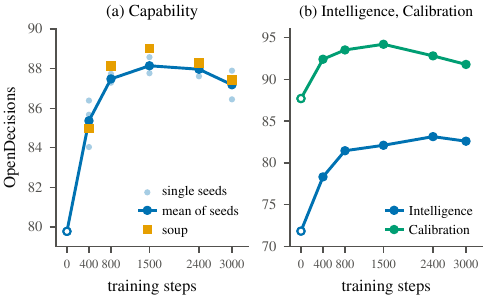}
  \caption{\textbf{Training length for Gemma~4~31B (Jeff~1.1 recipe).} (a) \jb\ Capability: mean of three seeds
  (line), single seeds (dots), and the soup of the three seeds (squares; at 1,500 steps: Jeff-1.1-Large). (b)
  Intelligence and Calibration, mean of three seeds. Step 0 (open circles): the zero-shot model. Every checkpoint ends
  with a 100-step learning-rate decay and gets
  its own temperatures, fitted as for the released models. Each step processes 65,536 padded tokens. Other models:
  Appendix~\ref{app:train}.}
  \label{fig:steps}
\end{figure}

We examine which parts of the training recipe improve performance
and which can be simplified. Table~\ref{tab:ablation} varies one
component of the Gemma~4~31B recipe of Jeff~1.0 at a time, using its three-seed
mean of 84.4 Capability as the reference. All variants use a
single run with seed 1. Reported intervals account for question
sampling and estimated training-seed variability, so small
differences should be interpreted cautiously. We study training length and attention with the Jeff~1.1 recipe instead (Figure~\ref{fig:steps} and Appendix~\ref{app:train}).

\paragraph{LoRA drives the improvement.}
Training only a decision head on frozen model representations
gives 78.9 Capability, no improvement over the calibrated
zero-shot score of 79.6. In contrast, training LoRA while keeping
the head fixed at its original answer-letter weights gives 84.9,
comparable to the 84.4 reference score from joint training.
The main benefit therefore comes from adapting the language
model, with no clear additional gain from training the head.

\paragraph{Preserving the original head initialization.}
We also test whether training the head before LoRA training provides a better starting point. We first train
a head on frozen representations, then use it to initialize
LoRA fine-tuning, either updating the head jointly or keeping
it fixed. These variants score 1.9--2.8 points below the
reference. Starting directly from the answer-letter slice of the pretrained LLM head avoids this additional training
stage and preserves zero-shot performance at LoRA initialization.

\paragraph{Starting from an instruction-tuned model.}
Our short fine-tuning procedure does not replace the benefits
of an instruction-tuned starting checkpoint. Applied to
pretrained-only Gemma~4~31B, the recipe reaches
70.3 Capability: 14.1 points below the instruction-tuned model
after fine-tuning, and 9.3 points below that model before
fine-tuning (see Appendix~\ref{app:readout} for more details).

\paragraph{Training helps up to about 1,500 steps.}
With the Jeff~1.1 recipe, Gemma~4~31B increases up to 1,500 training steps, and then decreases again (Figure~\ref{fig:steps}). Intelligence keeps rising
until 2,400 steps (78.3 after 400 steps, 83.1 after 2,400), while Calibration is highest at 1,500 steps and then falls (94.2 to 91.8 after 3,000
steps). Qwen3.6-27B and Gemma~4~12B also gain from 800 to 1,500 steps (\pmci{$+$1.7}{1.0} and \pmci{$+$1.8}{1.3}), and
Gemma~4~12B loses again at 2,300 steps. Qwen3.5-4B scores the same after 800 and 1,500 steps, so we train the released version for 800
(Appendix~\ref{app:train}).

\paragraph{Bidirectional attention makes little difference.} Surprisingly, we found bidirectional helps little, and even slightly hurts the 31B model (Capability drops by \pmci{$-$0.2}{1.1} at 1,500 steps, average over three seeds). Our released 31B uses causal attention, while we use bidirectional attention for the other three.

\paragraph{Stable across LoRA ranks.}
LoRA ranks 16, 32, 64, and 128 give almost identical Capability scores, indicating that a smaller LoRA rank could potentially make training cheaper without meaningful loss in performance; however, these results are only one seed each.

\paragraph{Teacher labels help, soft labels (possibly) don't.} Removing the approximately 50,000
questions labeled by language models gives 82.5, a decrease
of \pmci{1.9}{2.2} points. Replacing soft targets with hard labels (via argmax) produces no measurable
change in Capability.

\paragraph{Alternative encodings and training objectives.}
We also test an option encoding designed to make predictions
independent of option order, along with alternative losses for
the decision head (Appendix~\ref{app:negative}). The
order-invariant encoding provides no clear benefit on
Qwen3.5-4B. When training a head on frozen representations,
none of the tested losses clearly improves calibration after
temperature scaling. An expected-utility objective, which
computes the expected reward for correct, incorrect, and
abstained answers, improves accuracy by \pmci{0.9}{0.9}
percentage points over cross-entropy. Combined with LoRA on Gemma~4~31B, the expected-utility objective stays within the range of the recipe's three seeds (Appendix~\ref{app:negative}).

\section{Discussion and limitations}\label{sec:discussion}

\paragraph{LLMs are already decision models.}
Our results show that open-weight language models already achieve performance comparable to Jev \textit{without any additional training} and without the need for specialized architectures or calibrated reinforcement learning. We show that temperature scaling further
improves the calibration of these models. These findings reveal the latent capabilities of open-source LLMs for probabilistic classification.

\paragraph{Simple LoRA fine-tuning goes beyond Jev.}
We show that simple LoRA fine-tuning can improve the model's decision capabilities even further. On both \jb and the public JevBench leaderboard, our 4B model already roughly matches Jev, while our larger models exceed it.
Our ablations suggest that the main benefit comes from LoRA itself rather than learning a new decision head.
We show that LoRA fine-tuning with the
original head fixed performs comparably to joint training.
This suggests that the simple answer-letter readout is
already effective, and that improving the representations
it reads is more important in our setting. Jeff~1.1 differs from Jeff~1.0 mainly in its training data and length, and gains the most for the smaller models, which suggests that the training data matters more than the details of the method, and better training data could be the important knob for training better open decision models in the future.

\paragraph{Data.}
We open-source \jb, a validation dataset for decision models, and DecisionBank, our full training dataset.
While we aimed for diversity across tasks, difficulty, and languages in \jb, we note it may not represent all real-world applications accurately. The dataset is 78\% English, and many questions were generated by language models rather than written by real users. We also note that we selected some of our training recipe settings, such as training length, on \jb. We did not use \pub, the JevBench leaderboard, or the public decision and knowledge benchmarks to select the training recipes. Furthermore, we note that Qwen3.8-27B is our teacher model and also one of the three panel models that checked the labels of \jb' generated and real samples. Finally, our decision head has 16 answer slots, compared to Jev's 256 options. Our serving code answers questions with up to 256 options in groups of at most 16 and a final round between the group winners (up to 17 forward passes); we have not evaluated this mode systematically.

\paragraph{Future work.}
Although our findings reveal the latent capabilities of open-source models for probabilistic classification, this does not establish that reinforcement learning or specialized decision architectures cannot improve performance further. In fact, we believe that exploring novel architectures, training procedures, and RL methods to further improve performance can be an interesting avenue for future work. Another avenue of interest is exploring which training examples actually improve decision-making; while training data for LLMs is abundant, more high-quality data tailored towards decision models will likely help improve these models further.

\section*{Acknowledgments}
JG is
supported by a Kempner Graduate Fellowship. DAM acknowledges support from the Kempner Institute and NSF award no.~2229881.

\section*{AI Statement}
The authors used AI agents to help with writing code, running experiments, and writing. AI-generated content was thoroughly verified, and the authors take responsibility for the final contents of this work.

\let\url\nolinkurl
\bibliography{refs}
\bibliographystyle{icml2026}

\newpage
\appendix
\onecolumn

\section{Request format}\label{sec:requests}

A request has a state (text or a JSON object) and one or more named questions in TypeSafe's format. The example below has one question of each type. Each question is
answered in its own forward pass.

\begin{Verbatim}[frame=single, fontsize=\scriptsize]
{"state": "Order #4411 was charged twice. The second charge is still pending, and the customer asks
           for the money back.",
 "questions": {
   "team":       {"type": "choice", "instructions": "Which team should handle this ticket?",
                  "criteria": {"billing": "Payments, charges and invoices.",
                               "technical": "Bugs, errors and login problems.",
                               "refund": "Returns of purchased items."}},
   "money_owed": {"type": "noul", "instructions": "Is the customer owed money back?",
                  "criteria": {"true": "The customer was charged more than they should pay.",
                               "false": "Every charge is correct."}},
   "urgency":    {"type": "score", "instructions": "How urgent is this ticket?",
                  "criteria": ["Not urgent: can wait a week.", "Normal: handle within two days.",
                               "Urgent: handle today.", "Critical: handle now."]}}}
\end{Verbatim}

\paragraph{System message.} We adopt the system prompt of SemIf~\citep{semif}:

\begin{Verbatim}[frame=single, fontsize=\scriptsize]
Apply the supplied criterion to the supplied evidence. Choose exactly one listed option. Respond with only
its uppercase letter, with no explanation or reasoning.
\end{Verbatim}

\paragraph{User message.} One JSON object: the state (\texttt{evidence}), the question's instructions
(\texttt{criterion}) and the options, labeled A, B, C, and so on, each described as ``\texttt{<id>: <text>}''.
A \textit{choice question} lists its options in the order of the request. A \textit{yes/no question} becomes two options,
\texttt{true} and \texttt{false}. A \textit{score question} lists its levels from lowest to highest, with ids 0, 1, 2, and so on.
The three user messages of the example above (line breaks added for display; each message is one line):

\begin{Verbatim}[frame=single, fontsize=\scriptsize, label={\scriptsize choice}]
{"evidence": "Order #4411 was charged twice. The second charge is still pending, and the customer asks for
 the money back.", "criterion": "Which team should handle this ticket?", "options": [{"letter": "A",
 "description": "billing: Payments, charges and invoices."}, {"letter": "B", "description": "technical: Bugs,
 errors and login problems."}, {"letter": "C", "description": "refund: Returns of purchased items."}]}
\end{Verbatim}
\begin{Verbatim}[frame=single, fontsize=\scriptsize, label={\scriptsize yes/no}]
{"evidence": "Order #4411 was charged twice. The second charge is still pending, and the customer asks for
 the money back.", "criterion": "Is the customer owed money back?", "options": [{"letter": "A",
 "description": "true: The customer was charged more than they should pay."}, {"letter": "B",
 "description": "false: Every charge is correct."}]}
\end{Verbatim}
\begin{Verbatim}[frame=single, fontsize=\scriptsize, label={\scriptsize score}]
{"evidence": "Order #4411 was charged twice. The second charge is still pending, and the customer asks for
 the money back.", "criterion": "How urgent is this ticket?", "options": [{"letter": "A", "description":
 "0: Not urgent: can wait a week."}, {"letter": "B", "description": "1: Normal: handle within two days."},
 {"letter": "C", "description": "2: Urgent: handle today."}, {"letter": "D", "description":
 "3: Critical: handle now."}]}
\end{Verbatim}

\paragraph{Complete model input.} The system and user message are rendered with the model's chat template, with thinking
disabled: the template opens the model's turn and closes an empty thinking block, so the next token is the answer.
For Gemma 4, the choice question becomes this string of 145 tokens (user message as above, shortened for brevity here):

\begin{Verbatim}[frame=single, fontsize=\scriptsize, label={\scriptsize Gemma 4}]
<bos><|turn>system
Apply the supplied criterion to the supplied evidence. Choose exactly one listed option. Respond with only
its uppercase letter, with no explanation or reasoning.<turn|>
<|turn>user
{"evidence": "Order #4411 was charged twice. ...", "criterion": "Which team should handle this ticket?",
 "options": [...]}<turn|>
<|turn>model
<|channel>thought
<channel|>
\end{Verbatim}
\newpage
For the Qwen models, the input looks as follows:

\begin{Verbatim}[frame=single, fontsize=\scriptsize, label={\scriptsize Qwen3.6 and Qwen3.5}]
<|im_start|>system
Apply the supplied criterion to the supplied evidence. Choose exactly one listed option. Respond with only
its uppercase letter, with no explanation or reasoning.<|im_end|>
<|im_start|>user
{"evidence": "Order #4411 was charged twice. ...", "criterion": "Which team should handle this ticket?",
 "options": [...]}<|im_end|>
<|im_start|>assistant
<think>

</think>

\end{Verbatim}

\paragraph{Output.} The model reads the string once. The decision head (Eq.~\ref{eq:head}) maps the hidden state at
the last position to one logit per option slot; the first $K$ slots are used, and a softmax over them gives the
raw probabilities. Next, we divide the logits by the temperature of the question type and return one answer per
question. Table~\ref{tab:request-output} shows the numbers of \jeffL\ for the example (temperatures 1.52 for choice
and 1.31 for yes/no and score questions); the answer it returns is (rounded for display):

\begin{Verbatim}[frame=single, fontsize=\scriptsize]
{"team":       {"type": "choice", "choice": "billing",
                "probabilities": {"billing": 0.842, "technical": 0.010, "refund": 0.148}},
 "money_owed": {"type": "noul", "noul": 0.992},
 "urgency":    {"type": "score", "score": 2,
                "probabilities": {"0": 0.020, "1": 0.226, "2": 0.556, "3": 0.198}}}
\end{Verbatim}

A yes/no answer is the probability of \texttt{true}.

\begin{table}[h]
  \caption{\textbf{Output of \jeffL\ for the example request.} Head logit: the decision head's logit for the option
  (Gemma's logits are soft-capped at 30). Raw: softmax over the question's options. Served: after the temperature.}
  \label{tab:request-output}
  \centering\small
  \begin{tabular}{@{}lllccc@{}}
    \toprule
    Question & Letter & Option & Head logit & Raw & Served \\
    \midrule
    \texttt{team} (choice) & A & billing & 26.41 & 0.933 & \textbf{0.842} \\
     & B & technical & 19.65 & 0.001 & 0.010 \\
     & C & refund & 23.77 & 0.066 & 0.148 \\
    \midrule
    \texttt{money\_owed} (yes/no) & A & true & 24.13 & 0.998 & \textbf{0.992} \\
     & B & false & 17.74 & 0.002 & 0.008 \\
    \midrule
    \texttt{urgency} (score) & A & 0: not urgent & 23.03 & 0.008 & 0.020 \\
     & B & 1: normal & 26.22 & 0.194 & 0.226 \\
     & C & 2: urgent & 27.40 & 0.634 & \textbf{0.556} \\
     & D & 3: critical & 26.05 & 0.164 & 0.198 \\
    \bottomrule
  \end{tabular}
\end{table}

\paragraph{Training questions.} Training uses the same system message, user message and chat template, with two
differences in how the options of a training question are shown. A training question comes from a dataset (Table~\ref{tab:data})
and keeps that dataset's option ids and wording; a yes/no question keeps its own two options (for example ``yes,
entailment'' and ``no, not entailment''). In each epoch the options are shown in a new random order; score levels keep
their order. For Qwen3.5-4B, a fixed half of the training questions is instead shown in the evaluation format above:
a yes/no question becomes the options \texttt{true} and \texttt{false}, \texttt{true} first, and a choice question lists
its options in one fixed order (sorted by id, since a training question has no request order). The other models are
trained on the first format only. The options of one yes/no training question in both formats (texts shortened):

\begin{Verbatim}[frame=single, fontsize=\scriptsize, label={\scriptsize training format (one epoch)}]
"options": [{"letter": "A", "description": "no, not entailment: Text 1 does not support Text 2 ..."},
            {"letter": "B", "description": "yes, entailment: Text 1 logically guarantees Text 2 ..."}]
\end{Verbatim}
\begin{Verbatim}[frame=single, fontsize=\scriptsize, label={\scriptsize evaluation format}]
"options": [{"letter": "A", "description": "true: entailment: Text 1 logically guarantees Text 2 ..."},
            {"letter": "B", "description": "false: not entailment: Text 1 does not support Text 2 ..."}]
\end{Verbatim}

\section{Training details}\label{app:train}

\begin{table}[h]
  \caption{Training settings of Jeff~1.1.}
  \centering\small
  \begin{tabular}{@{}ll@{}}
    \toprule
    Setting & Value \\
    \midrule
    LoRA & rank 32, $\alpha = 64$, dropout 0; q/k/v/o, gate/up/down projections (Qwen: also in\_proj\_qkv, in\_proj\_z, out\_proj) \\
    Head & decision head (Eq.~\ref{eq:head}), 16 slots, initialized from the language-model head, trained jointly \\
    Optimizer & AdamW, LR $2 \cdot 10^{-5}$ (LoRA), $10^{-4}$ (head); Qwen3.5-4B: $10^{-4}$ for both; weight decay 0 \\
    Schedule & WSD schedule with 250 warmup steps, and decay over the last 100 steps; 1,500 steps total (Qwen3.5-4B: 800) \\
    Batch & 65,536 padded tokens per step \\
    Inputs & at most 8,192 tokens; options shuffled; bf16 \\
    Attention & bidirectional over the prompt, except Gemma 4 31B (causal) \\
    Loss & cross-entropy against the soft target \\
    Runs & three seeds per model; the released model is the model soup (average) of their weights \\
    Calibration & per-type temperatures (at least 1) fitted by NLL on held-out task families; yes/no rule (Section~\ref{sec:method}) \\
    \bottomrule
  \end{tabular}
\end{table}

\paragraph{The Jeff 1.0 recipe.} The ablations of Section~\ref{sec:ablations} use the recipe of our 1.0 suite of models: the same LoRA and head on a pool of 417,825 questions (Appendix~\ref{app:jeff10}), 400 steps (250 warmup steps, constant to step 300, cosine decay to step 400), three runs reported as their mean, and causal attention.

\begin{figure}[h]
  \centering
  \includegraphics[width=\textwidth]{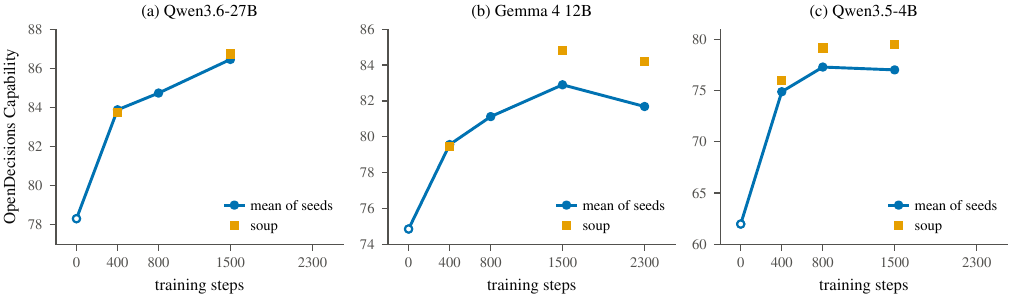}
  \caption{\textbf{Training length for the 27B, 12B, and 4B models (Jeff~1.1 recipe).} \jb\ Capability, mean of three seeds and the soup of the three seeds. Released models are soups at 1,500 steps for the 27B and 12B, and at 800 steps for the 4B model. Step 0 (open
  circles) is the zero-shot model.}
  \label{fig:steps-small}
\end{figure}

\paragraph{Training length.} Figures~\ref{fig:steps}
and~\ref{fig:steps-small} use the Jeff~1.1 recipe. Every checkpoint ends with a 100-step learning-rate decay; for example, the
400-step runs decay from the step-300 checkpoint. Every evaluated checkpoint gets its own per-type temperatures, fitted on the held-out data of the released
models in the same way.

\section{Jeff 1.0 and Jeff 1.1}\label{app:jeff10}

Our first release, Jeff~1.0, used the same architecture, loss, and LoRA settings as Jeff~1.1. It differs in five ways:
the training data (the pool of Table~\ref{tab:pool10}, with teacher labels from Qwen3.5-122B-A10B and Qwen3.6-27B, instead of
DecisionBank), 400 instead of 1,500 training steps (as more steps were not beneficial on the previous training pool), no model souping, causal attention for
every model, and temperatures fitted on held-out questions from the training pool. For Qwen3.5-4B, Jeff~1.0 also used a LoRA
learning rate of $10^{-4}$, distillation from the trained 31B model, and half of the training questions in the evaluation
format. Table~\ref{tab:jeff10} compares the 1.0 and 1.1 model suites on \jb and \pub. Jeff~1.1 is better for every model, by between 4.1 points
for the 31B and 12.6 points for the 4B model.

\begin{table}[h]
  \caption{\textbf{Comparison of Jeff 1.0 and Jeff 1.1} on \jb\ (Intelligence, Calibration, and Capability),
  the difference in Capability (\pmci{}{1.96 SE}, question bootstrap), and \pub\ Capability.}
  \label{tab:jeff10}
  \centering\small
  \begin{tabular}{@{}lccccc@{}}
    \toprule
    & \multicolumn{2}{c}{\jb\ $I$ / $C$ / Capability} & $\Delta$ Capability & \multicolumn{2}{c}{\pub} \\
    \cmidrule(lr){2-3}\cmidrule(lr){5-6}
    Model & Jeff 1.0 & Jeff 1.1 & & Jeff 1.0 & Jeff 1.1 \\
    \midrule
    Large (Gemma 4 31B) & 76.8 / 92.9 / 84.9 & 83.5 / 94.5 / 89.0 & \pmci{$+$4.1}{1.7} & 81.0 & 82.2 \\
    Qwen-27B (Qwen3.6-27B) & 73.8 / 89.0 / 81.4 & 81.2 / 92.5 / 86.8 & \pmci{$+$5.4}{2.1} & 81.3 & 83.6 \\
    Medium (Gemma 4 12B) & 65.2 / 83.9 / 74.5 & 77.2 / 92.8 / 85.0 & \pmci{$+$10.4}{2.6} & 78.4 & 83.2 \\
    Qwen-4B (Qwen3.5-4B) & 54.1 / 79.2 / 66.7 & 68.4 / 90.2 / 79.3 & \pmci{$+$12.6}{2.8} & 63.7 & 72.7 \\
    \bottomrule
  \end{tabular}
\end{table}

\begin{table}[h]
  \caption{\textbf{Training pool of Jeff 1.0} (417,825 questions). Soft = the target is a distribution (annotator votes or
  teacher probabilities). Teacher labels come from Qwen3.5-122B-A10B and Qwen3.6-27B.}
  \label{tab:pool10}
  \centering\small
  \begin{tabular}{@{}lrl@{}}
    \toprule
    Part & Rows & Labels \\
    \midrule
    decider public mixture (79 tasks) & 300,042 & dataset (hard) \\
    Civil Comments & 30,000 & annotators (soft) \\
    Measuring Hate Speech & 30,000 & annotators (soft) \\
    STS-B & 4,760 & similarity (soft) \\
    ChaosNLI & 2,886 & 100 votes (soft) \\
    synthetic documents & 30,641 & teachers (soft) \\
    decider custom questions & 19,496 & teachers (soft) \\
    \midrule
    Total & 417,825 & 117,783 soft \\
    \bottomrule
  \end{tabular}
\end{table}

\section{\jb\ details}\label{app:jb16}

\begin{table}[h]
  \caption{\jb\ composition by question type and tier (target in brackets: the composition of the public JevBench v1.6
  split times 8).}
  \centering\small
  \begin{tabular}{@{}lrrrrr@{}}
    \toprule
    Type & Easy & Standard & Judge & Hard & Total \\
    \midrule
    Choice & 261 (264) & 136 (136) & 255 (256) & 526 (544) & 1,178 (1,200) \\
    Yes/no & 110 (112) & 144 (144) & 158 (160) & 174 (184) & 586 (600) \\
    Score & 72 (72) & 127 (128) & 168 (168) & 221 (232) & 588 (600) \\
    \bottomrule
  \end{tabular}
\end{table}

\paragraph{Agreement with the public leaderboard.} We ran 17 public systems, including the Jev API, as their authors
serve them and compared their scores, under the leaderboard's own scorer, with their official v1.6.1 Capability
(Table~\ref{tab:agree}). For the 10 systems with official Capability of at least 55 that were also used while building
\jb, \jb\ ranks them in nearly the official order (Spearman 0.98). Four systems were never used while building the set
(Eikos-27B, Clef, AutoJev-27B and Open-Jev-27B; official Capability 68.0 to 77.2). A line fitted on the 10 old systems
predicts their official scores with a root mean square error of 3.0 points (95\% interval 2.2 to 3.9) and a mean error
of $+0.5$; 78\% of new-old pairs come out in the official order. Over all 14 systems the Spearman correlation is 0.90.
\pub\ predicts worse (error 5.4), and our training-distribution validation set fails (16.2).

\begin{table}[h]
  \caption{\textbf{Does a set predict the official leaderboard?} Agreement between Capability on each set and the
  official v1.6.1 Capability of public systems with official Capability $\geq 55$, each run as its authors serve it and
  scored with the leaderboard's scorer. LOO = mean absolute error when predicting each system's official score from the
  others by a linear fit. Gray brackets: 95\% intervals from resampling item groups (correlations are bounded, so we
  show the interval itself rather than $\pm$).}
  \label{tab:agree}
  \centering\small
  \begin{tabular}{@{}lrlcr@{}}
    \toprule
    Set & Items & Spearman & Pairs in order & LOO \\
    \midrule
    \multicolumn{5}{@{}l}{\emph{10 systems used while building \jb}} \\
    \pub & 231 & \ciint{0.85}{0.56}{0.94} & 0.84 & 4.9 \\
    \jb\ pilot & 283 & \ciint{0.90}{0.67}{0.92} & 0.89 & 4.3 \\
    \jb & 2,352 & \ciint{0.98}{0.92}{0.99} & 0.96 & 2.6 \\
    Training-distribution validation set & 3,884 & \ciint{0.45}{0.31}{0.53} & 0.67 & 10.9 \\
    \midrule
    \multicolumn{5}{@{}l}{\emph{Adding 4 systems not used to build \jb}} \\
    \jb, new systems only & 2,352 & 0.80 & 0.83 & 4.1 \\
    \jb, all 14 systems & 2,352 & \ciint{0.90}{0.88}{0.95} & 0.87 & 2.7 \\
    \bottomrule
  \end{tabular}
\end{table}

\paragraph{Results per family.} The families of \jb\ separate systems in different ways (Table~\ref{tab:family}). StartLux-27B Instant is best on six of the eleven families, and Jeff-1.1-Large or Jeff-1.1-Qwen-27B on counting traps, composite rules, the generated calibration items, and option order (tied with StartLux-27B Instant). Jev leads on the calibration items, and every system gets
the negation pairs right.

\begin{table}[h]
  \caption{\textbf{Competence by item family on \jb} (chance-corrected accuracy; same items for all systems; zero-shot with
  temperatures; public systems as in Table~\ref{tab:main}).}
  \label{tab:family}
  \centering\small
  \setlength{\tabcolsep}{3.5pt}
  \resizebox{\textwidth}{!}{%
  \begin{tabular}{@{}lccccccccc@{}}
    \toprule
    Family & Jeff-1.1- & Gemma 4 31B & Jeff-1.1- & Qwen3.6-27B & StartLux-27B & pplx-decider- & Quyet-L & deck31b & Jev \\
    & Large & zero-shot & Qwen-27B & zero-shot & Instant & v1.3 & & & \\
    \midrule
    LLM routing & 82.0 & 76.1 & 81.9 & 76.6 & \textbf{84.6} & 80.7 & 77.7 & 76.5 & 79.3 \\
    Multilingual & 84.0 & 76.8 & 79.5 & 72.1 & \textbf{87.9} & 79.4 & 76.6 & 78.0 & 67.5 \\
    Fairness pairs & 84.0 & 75.4 & 79.3 & 60.1 & \textbf{94.2} & 85.5 & 86.6 & 76.9 & 72.0 \\
    Use cases & 81.3 & 68.6 & 74.2 & 59.1 & \textbf{84.1} & 75.2 & 71.6 & 68.1 & 54.4 \\
    Counting traps & \textbf{76.1} & 53.4 & 73.2 & 34.5 & 65.2 & 59.7 & 57.4 & 56.8 & 60.4 \\
    Negation pairs & 100 & 100 & 100 & 100 & 100 & 100 & 100 & 100 & 100 \\
    Option order & 89.6 & 85.5 & \textbf{90.7} & 84.4 & \textbf{90.7} & 89.6 & 84.4 & 85.5 & 85.5 \\
    Perturbation pairs & 74.5 & 55.5 & 75.6 & 58.4 & \textbf{87.9} & 79.3 & 68.0 & 63.1 & 68.3 \\
    Composite rules & 96.7 & 79.3 & \textbf{99.2} & 64.4 & 85.1 & 92.9 & 85.6 & 77.1 & 76.0 \\
    Calibration (generated) & \textbf{91.7} & 65.7 & 91.5 & 72.8 & 87.4 & 81.1 & 83.1 & 74.9 & 77.3 \\
    Calibration & 84.6 & 57.8 & 80.7 & 63.3 & 84.0 & 81.5 & 83.0 & 73.2 & \textbf{94.1} \\
    \bottomrule
  \end{tabular}
  }
\end{table}

\subsection{Training--evaluation overlap and distribution differences}
\label{app:train-eval}

\paragraph{DecisionBank.} We filtered DecisionBank against the evaluation sets when we built it: no question has the same state (the sample's context, compare Appendix~\ref{sec:requests}) as an \jb\ or \pub\ item, and we removed teacher-written questions that share at least two word 8-grams with an evaluation item. The 28,442 programmatic questions come from the same generator code as the 599 programmatic \jb\ items, with new random seeds. Trained without them (three runs of 1,500 steps and their average; temperatures fitted on held-out training questions), Gemma~4~31B Capability on \jb drops by 1.8. The rest of this section compares \jb\ with DecisionBank.

 We compare \jb\ with the 261,561 questions of DecisionBank. This analysis uses an earlier draw of \jb\ with
2,389 questions, which contains all 2,352 final questions. We examine whether evaluation questions
appear in the training data and how similar the two datasets are.
The comparisons use the stored questions, supporting context,
and answer options, rather than the exact prompts seen by the model.

\paragraph{Exact overlap.}
We check for exact matches between training and evaluation using
the complete question inputs and the supporting
context (state) alone. We repeat these checks after standardizing
whitespace and Unicode. We find no exact matches in any of the
2,389 evaluation questions. This rules out direct reuse of
identical inputs under these checks, but does not rule out
paraphrased questions or shared underlying skills.

\paragraph{Similarity to training data.}
We also measure how closely evaluation questions resemble
training questions. We hold out 2,389 training questions ($H$)
and use the remaining 259,172 ($R$) as a reference set.
Using TF--IDF features and cosine similarity, we find the
closest question in $R$ for each question in $H$ and \jb.
For this comparison, we cut every input to 8,000 characters, which shortens
the 37 longest evaluation contexts; the exact-match
checks above use the full text.

The median similarity is 0.240 for \jb, compared with 0.555
for held-out training questions (Table~\ref{tab:train-eval-similarity}).
Part of the training similarity comes from repeated supporting
contexts. After excluding training questions whose context
appears exactly in $R$, the median falls to 0.440, still
substantially above that of \jb. The difference also persists
when comparisons are restricted to the same question type.

\begin{table}[htbp]
\centering
\caption{Median nearest-neighbor TF--IDF cosine similarity
to the same training reference set $R$. The last row excludes
training questions with an identical supporting context in $R$.}
\label{tab:train-eval-similarity}
\begin{tabular}{lrr}
\toprule
Query set & Questions & Median similarity \\
\midrule
\jb & 2,389 & 0.240 \\
Held-out training ($H$) & 2,389 & 0.555 \\
$H$ without shared contexts & 1,962 & 0.440 \\
\bottomrule
\end{tabular}
\end{table}

\paragraph{Distinguishing the datasets.}
As a separate check, we train a logistic-regression classifier
using TF--IDF features to predict whether a question comes
from training or evaluation. We use 2,389 questions from each
dataset, grouped five-fold cross-validation, and three random
seeds. Related examples are kept in the same fold to reduce
leakage, and features are fitted only on the training folds. The classifier achieves 90.82\% balanced accuracy, with a
standard deviation of 0.11 percentage points across seeds.
This shows that the datasets differ in observable input text,
although differences in wording, formatting, and task
composition may contribute to the result.

\paragraph{Interpretation.}
We find no exact matches between \jb\ and our training pool, and the two datasets differ substantially in their input distributions. However, this does not mean that the evaluation questions require new skills or are necessarily harder. These differences also do not explain why models perform better or worse. We cannot rule out paraphrased overlap or overlap with the backbones' pretraining data. Our analysis covers the exact training data of the Jeff~1.1 models.

\section{Public evaluation sets}\label{app:public}

\begin{table}[h]
  \caption{\textbf{Official JevBench leaderboard}~\citep{jevbench}: the open systems with the highest Capability, the best
  open systems with at most 12B and at most 4B parameters, and our models, as published in November 2026 (revision v1.8.10).
  Each system is measured on its own draw of items. $^*$Evaluated by the leaderboard, not yet officially ranked. Jev is
  listed as a reference.}
  \label{tab:board}
  \centering\small
  \begin{tabular}{@{}lccc@{}}
    \toprule
    System & $I$ & $C$ & Capability \\
    \midrule
    Torchcast-decision-31B & 85.5 & 93.4 & 89.5 \\
    MetaDecision-31B-rain & 85.0 & 91.5 & 88.3 \\
    Karpo-decision-31B & 80.2 & 94.1 & 87.2 \\
    Jeff-1.1-Large (ours)$^*$ & 81.6 & 90.1 & 85.9 \\
    Mammoth-1 & 76.2 & 94.3 & 85.2 \\
    Jeff-1.1-Medium (ours) & 74.5 & 89.3 & 81.9 \\
    Torchcast Decision 4B (v1.3) & 67.1 & 87.8 & 77.5 \\
    Arcbright Elf 1 12B (r2) & 63.9 & 91.1 & 77.5 \\
    Jeff-1.1-Qwen-4B (ours) & 65.3 & 89.2 & 77.3 \\
    Jev 1.13.0 (API) & 63.6 & 90.6 & 77.1 \\
    \bottomrule
  \end{tabular}
\end{table}

\paragraph{Overlap with our training data.} We drop an evaluation item when it shares a 13-word sequence with our
training data (DecisionBank), unless the shared text is only template wording that a task repeats across its items. This
rule applies to the decision sets and to the knowledge benchmarks below.

\paragraph{Held-out decision datasets.} TREC (500 items), CB (56) and WANLI (256, SemIf's sample) are public
classification and inference sets; SemIf's two test sets (LLM-generated, 144 items; perturbations, 108) are public decision
items with paired variants; TypeSafe's example set (348 items) has reference answers written by strong LLMs, so we
report agreement with the most likely reference answer; decider's 29 held-out tasks (7,800 items) are built from public
datasets that decider keeps out of its training mix. The first six sets share no
13-word sequence with our training data, and the 52 decider items that do (71 against the Jeff~1.0 pool) share only
template wording, so we keep them. Inputs up
to 16,384 tokens are read in full by our evaluation code, which runs the models directly. Our serving code reads at most 8,192 tokens and shortens longer inputs; this concerns 212 of the 9,212 items (125 TypeSafe and 87 decider items).

\paragraph{Knowledge-heavy multiple choice.} We draw seeded samples of the sizes used by \citet{li2026llmasjev}: BoolQ
(500), MMLU (1,200), MMLU-Pro (800, 10 options), ARC-Challenge (600), WinoGrande (500) and SciQ (400), from splits we did
not train on. Our training pool contains the training splits of BoolQ, ARC, MMLU (auxiliary\_train) and WinoGrande, so
for our trained models these are seen tasks with unseen items; MMLU-Pro and SciQ are not. We drop 85
items due to overlaps with our training data, as described above.

\begin{table}[h]
  \caption{\textbf{Knowledge-heavy multiple-choice benchmarks}: accuracy (\%) on our random samples (fixed seed) of each benchmark. ECE with 15 bins. Mean over tasks. For reference, \citet{li2026llmasjev}
  report average accuracies on their own samples of 78.6 (4B, training-free), 78.9 (4B, LoRA), 82.2 (Winnow-12B), 75.4
  (SemIf), and 91.5 (GPT-5.6 Sol, text output).}
  \label{tab:external}
  \centering\small
  \setlength{\tabcolsep}{3.5pt}
  \begin{tabular}{@{}lcccccccc@{}}
    \toprule
    System & BoolQ & MMLU & MMLU-Pro & ARC-C & WinoGr. & SciQ & Average & ECE \\
    \midrule
    Jev 1.13.0 (API) & \textbf{92.6} & \textbf{91.8} & \textbf{84.0} & \textbf{97.8} & \textbf{92.2} & 98.7 & \textbf{92.9} & .032 \\
    Quyet-1.0-Large & 92.1 & 85.3 & 65.9 & 97.5 & 90.0 & \textbf{99.0} & 88.3 & .035 \\
    deck31b & 92.3 & 85.8 & 68.8 & 97.2 & 86.0 & 98.7 & 88.1 & .077 \\
    decider-4b-v2 & 91.4 & 75.0 & 52.3 & 92.7 & 82.8 & 98.2 & 82.1 & .038 \\
    Torchcast-decision-12B & 89.6 & 79.0 & 52.8 & 94.3 & 75.8 & 98.0 & 81.6 & .050 \\
    Winnow-12B & 88.9 & 77.3 & 54.0 & 94.7 & 74.6 & 98.0 & 81.2 & .084 \\
    Plumb-4B & 87.9 & 73.5 & 53.4 & 92.5 & 68.0 & 98.0 & 78.9 & .073 \\
    SemIf & 86.5 & 69.2 & 40.5 & 89.3 & 60.0 & 98.0 & 73.9 & .083 \\
    \midrule
    Gemma 4 31B, zero-shot & \textbf{92.6} & 85.9 & 70.3 & 96.8 & 85.0 & 98.7 & 88.2 & .110 \\
    Jeff-1.1-Large & 92.3 & 87.9 & 73.2 & 97.5 & 87.6 & 98.7 & 89.5 & .035 \\
    Qwen3.6-27B, zero-shot & 91.6 & 84.8 & 62.8 & 96.5 & 84.6 & 98.5 & 86.5 & .054 \\
    Jeff-1.1-Qwen-27B & 91.4 & 88.4 & 71.5 & 97.2 & 90.2 & 98.7 & 89.6 & .033 \\
    Gemma 4 12B, zero-shot & 88.4 & 75.3 & 54.8 & 93.5 & 69.2 & 98.5 & 79.9 & .167 \\
    Jeff-1.1-Medium & 90.3 & 79.9 & 62.8 & 96.2 & 81.0 & 98.5 & 84.8 & .041 \\
    Qwen3.5-4B, zero-shot & 89.1 & 69.8 & 41.3 & 90.3 & 63.4 & 98.0 & 75.3 & .079 \\
    Jeff-1.1-Qwen-4B & 89.3 & 78.1 & 56.6 & 95.2 & 77.8 & 98.7 & 82.6 & .039 \\
    \bottomrule
  \end{tabular}
\end{table}

\section{Zero-shot accuracy and model size}\label{app:scaling}

\begin{figure}[h]
  \centering
  \includegraphics[width=.7\columnwidth]{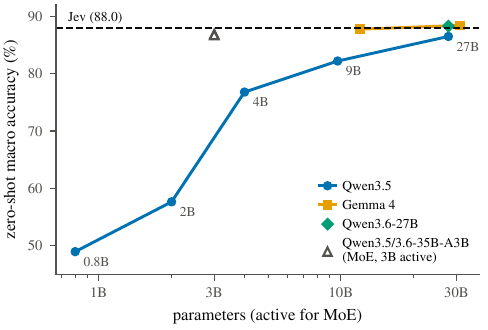}
  \caption{\textbf{Zero-shot accuracy against model size}: average accuracy (\%) over the seven held-out decision sets of
  Table~\ref{tab:heldout}, letter readout, no training. Mixture-of-experts models are placed at their 3B active
  parameters. Dashed: Jev.}
  \label{fig:scaling}
\end{figure}

Figure~\ref{fig:scaling} shows the zero-shot letter readout of eight Qwen models and two Gemma models on the seven
held-out decision sets of Table~\ref{tab:heldout}, scored in the same way (average accuracy on the same 9,212 items).
Temperatures do not change which option is most likely, so nothing is fitted here. Accuracy rises steeply up to 4B
parameters and then levels off: Qwen3.5 scores 48.9 at 0.8B, 57.6 at 2B, 76.8 at 4B, 82.2 at 9B and 86.5 at 27B. From
about 12B on, the model family matters as much as the size. Gemma~4~12B (87.8) is above Qwen3.5-27B (86.5), and
Gemma~4~31B (88.4), Qwen3.6-27B (88.3) and Jev (88.0) are within half a point of each other. The two mixture-of-experts
models, with 35B parameters of which 3B are active per token, score 86.9 (Qwen3.5) and 86.7 (Qwen3.6), close to the
dense 27B model at a fraction of its compute per token. For the Qwen models other than Qwen3.6-27B and Qwen3.5-4B we use
earlier zero-shot runs with the same prompt and readout; for these two models, the earlier runs and the runs of
Table~\ref{tab:heldout} agree to within 0.1 points.

\section{Order-invariant encoding and calibration losses}\label{app:negative}

Before we settled on the recipe, we tested two other ideas on Qwen3.5-4B. Both studies use the same training pool
(Table~\ref{tab:pool10}) but an earlier training setup, so the numbers are not directly comparable to the main text. We
rescore all runs on the seven held-out decision sets of Table~\ref{tab:heldout} (Table~\ref{tab:negative}). Intervals
include seed variation as in the main text: within each study we pool the seed variance of the variants that have
several seeds, and a single run counts it fully.

\paragraph{Order-invariant encoding.} With the letter readout, the answer can depend on the order of the options. After
LoRA training, the 4B model still changes its answer on about 6\% of choice and yes/no questions when the option order
is reversed. We therefore built an encoding that is order-invariant by construction. The state and question are encoded
once. Each option is then appended in its own short segment, all segments get the same positions, and no option can see
the others. The score of option $j$ is read from the final hidden state $h_j$ of its segment as
$s_j = w_0^\top h_j + a^\top \mathrm{std}(h_j) + b$, where $w_0$ is the difference of the language-model-head rows for
``yes'' and ``no'', and $a$, $b$ start at zero, as in the decision head (Eq.~\eqref{eq:head}). The probabilities are the
softmax of the scores over the options. We tried three variants: a set layer (two attention layers without positions)
that lets the option vectors interact; a clearer wording for each option segment (\texttt{Is the answer "<option>"?});
and the same wording with all options, sorted by text, listed once in the shared part. Each run trains LoRA (rank 16 on
all projections, learning rate $10^{-4}$) and the head (learning rate $10^{-3}$) for 600 steps of 262,144 tokens, on
the same batches.

All order-invariant variants score below the letter format, by 0.6 to 2.4 points. With one run per variant these
differences are within noise, so we can only say that the encoding does not help. The losses are largest on SemIf's
perturbation set and on TypeSafe's set, where the options must be compared with each other. Listing all options in the
shared part removes most of the gap, but then the encoding is close to a letter prompt without the letters. A simpler
change gives the same invariance at no cost: sort the options by text before building the letter prompt. The prompt is
then the same for every input order of a choice or yes/no question, and accuracy does not change (+1.0 points when we
sort only at test time, $-0.4$ when we also sort in training; both within noise). No variant changes the ECE by more
than 0.013. We therefore keep the letter format; sorting the options is a free option when exact invariance is needed.

\paragraph{Calibration losses.} We trained a head on the frozen 4B model with different losses. The head is a linear
map of the final hidden state $h$, initialized with the letter rows $W_0$ (Eq.~\eqref{eq:readout}) and trained on
cached hidden states (AdamW, learning rate $10^{-3}$, batch 8,192, at most 3,000 steps, early stopping on the
validation loss). Let $p$ be the predicted distribution over the $K$ options and $y$ the correct option. Besides
cross-entropy $-\log p_y$, we tried the Brier score $\sum_j (p_j - \mathbf{1}[j = y])^2$~\citep{brier1950verification};
focal loss $-(1 - p_y)^{\gamma} \log p_y$ with $\gamma = 2$~\citep{mukhoti2020focal}; cross-entropy plus $\lambda$ times
a differentiable ECE with soft bin memberships~\citep{karandikar2021soft} or plus $\lambda$ times MMCE, a kernel measure
of calibration error~\citep{kumar2018mmce}; cross-entropy against a target that mixes the label with the zero-shot
distribution, with weight $\alpha$ on the label; and an expected-utility objective. Rows with a soft target use it in
place of the label.

\paragraph{Expected utility.} This objective is our closest analog to TypeSafe's reinforcement learning for calibrated
decisions. The head gets one extra output, ``abstain''. Let $\pi$ be the softmax over the $K$ options and abstain. A
correct answer earns $+1$, a wrong answer $-r$, and abstaining $0$. The loss is the negative expected reward,
\begin{equation}\label{eq:eu}
  \mathcal{L}_{\text{EU}} = -\Big(\pi_y - r \sum_{j \le K,\, j \neq y} \pi_j\Big),
\end{equation}
averaged with cross-entropy on the $K$ options (weight $1/2$ each). Policy-gradient reinforcement learning with this
reward maximizes the same quantity in expectation; with a known label we compute it exactly and need no sampling. The
best policy answers only when its top option has probability at least $r/(1+r)$, so the objective rewards knowing when
it is unsure. At test time the abstain output is dropped and the option probabilities are renormalized. We use $r = 1$
and $r = 3$.

\paragraph{Results.} After temperature scaling, no loss lowers the mean ECE by more than 0.01 compared with
cross-entropy (0.092). Focal loss is clearly worse in accuracy (\pmci{$-$2.1}{1.7} points), and most other losses change
accuracy by less than one point. The expected-utility objective with $r = 3$ does best: over three seeds it is
\pmci{+0.9}{0.9} points above cross-entropy (79.4 against 78.5). Its raw ECE is lower (0.076 against 0.092), but after
temperature scaling it is not (0.094). With the decision-head form of Eq.~\eqref{eq:head}, the same objective gives
\pmci{+0.5}{1.4} points. On the frozen Qwen3.6-27B, no head beats the zero-shot readout in accuracy (88.3), with or
without the expected-utility objective. All of these heads stay far below LoRA training: the recipe raises the 4B model
to 85.0 on these sets (Table~\ref{tab:heldout}). Combined with LoRA on Gemma~4~31B (400 steps, seed 1), the
expected-utility objective mixed with cross-entropy scores 83.7 ($r = 1$) and 84.8 ($r = 3$) on \jb, within the range
of the recipe's three seeds (83.8 to 85.3), and 81.2 without cross-entropy. It is not needed to reach and exceed Jev.

\begin{table}[h]
  \caption{\textbf{Order-invariant encoding and calibration losses}, rescored on the seven held-out decision sets of
  Table~\ref{tab:heldout}. Acc.: average accuracy (\%). $\Delta$: difference to the first row of the block. ECE:
  mean over the six hard-label sets, raw and with one temperature fitted on the run's own validation predictions (+T).
  Rows with several seeds show the seed mean.}
  \label{tab:negative}
  \centering\small
  \begin{tabular}{@{}lccccc@{}}
    \toprule
    Variant & Seeds & Acc. & $\Delta$ & ECE & ECE +T \\
    \midrule
    \multicolumn{6}{@{}l}{\emph{Qwen3.5-4B, option encoding (LoRA and head, 600 steps)}} \\
    Letters, given order & 2 & \pmci{85.4}{2.6} & -- & .100 & .096 \\
    Letters, sorted at test time & 1 & \pmci{86.3}{3.3} & \pmci{+1.0}{3.5} & .103 & .096 \\
    Letters, sorted in training and test & 2 & \pmci{84.9}{2.6} & \pmci{$-$0.4}{2.9} & .092 & .089 \\
    Option-parallel & 1 & \pmci{83.4}{3.6} & \pmci{$-$1.9}{3.8} & .092 & .092 \\
    \quad + set layer & 1 & \pmci{83.1}{3.6} & \pmci{$-$2.3}{3.8} & .102 & .102 \\
    \quad clearer wording & 1 & \pmci{83.0}{3.6} & \pmci{$-$2.4}{3.7} & .085 & .083 \\
    \quad clearer wording + sorted option list & 1 & \pmci{84.8}{3.2} & \pmci{$-$0.6}{3.6} & .091 & .087 \\
    \midrule
    \multicolumn{6}{@{}l}{\emph{Qwen3.5-4B, loss for a head on the frozen model}} \\
    Cross-entropy & 3 & \pmci{78.5}{2.5} & -- & .092 & .092 \\
    Zero-shot (no training) & 1 & \pmci{76.8}{2.8} & \pmci{$-$1.7}{2.0} & .109 & .096 \\
    Brier score & 1 & \pmci{79.6}{2.6} & \pmci{+1.1}{1.3} & .093 & .094 \\
    Focal loss, $\gamma = 2$ & 1 & \pmci{76.3}{2.7} & \pmci{$-$2.1}{1.7} & .193 & .099 \\
    CE + soft-binned ECE, $\lambda = 0.1$ & 1 & \pmci{78.4}{2.6} & \pmci{0.0}{0.6} & .093 & .095 \\
    \quad $\lambda = 0.3$ & 3 & \pmci{78.5}{2.6} & \pmci{0.0}{0.5} & .089 & .090 \\
    \quad $\lambda = 1$ & 1 & \pmci{78.3}{2.7} & \pmci{$-$0.2}{1.6} & .102 & .097 \\
    \quad $\lambda = 3$ & 1 & \pmci{78.4}{2.5} & \pmci{0.0}{1.6} & .113 & .118 \\
    CE + MMCE, $\lambda = 0.5$ & 1 & \pmci{78.6}{2.6} & \pmci{+0.1}{0.6} & .090 & .091 \\
    \quad $\lambda = 2$ & 1 & \pmci{78.8}{2.6} & \pmci{+0.3}{0.6} & .086 & .086 \\
    CE to mixed target, $\alpha = 0.5$ & 1 & \pmci{78.4}{2.7} & \pmci{$-$0.1}{1.5} & .083 & .082 \\
    \quad $\alpha = 0.8$ & 1 & \pmci{79.1}{2.5} & \pmci{+0.6}{1.0} & .088 & .088 \\
    Expected utility, $r = 1$ & 1 & \pmci{79.1}{2.7} & \pmci{+0.6}{1.2} & .094 & .094 \\
    \quad $r = 3$ & 3 & \pmci{79.4}{2.6} & \pmci{+0.9}{0.9} & .076 & .094 \\
    \midrule
    \multicolumn{6}{@{}l}{\emph{Qwen3.6-27B, head on the frozen model}} \\
    Zero-shot (no training) & 1 & \pmci{88.3}{1.8} & -- & .065 & .089 \\
    Decision head, cross-entropy & 1 & \pmci{87.7}{1.8} & \pmci{$-$0.6}{0.9} & .079 & .081 \\
    Decision head, mixed target, $\alpha = 0.5$ & 1 & \pmci{88.1}{1.8} & \pmci{$-$0.2}{0.7} & .056 & .071 \\
    Linear head, expected utility, $r = 3$ & 1 & \pmci{88.2}{1.8} & \pmci{$-$0.2}{0.9} & .058 & .063 \\
    \bottomrule
  \end{tabular}
\end{table}

\section{Calibration}\label{app:calib}

This appendix concerns temperatures fitted on held-out questions from the training data, as used for the zero-shot and trained models. This section mostly considers the Jeff-1.0 recipe; Table~\ref{tab:temps11} lists the temperatures for the Jeff-1.1 models.

Table~\ref{tab:temps} compares Capability scores before and after temperature scaling, while Table~\ref{tab:tempvalues} reports the fitted temperatures for each question type. The largest improvements occur for the zero-shot Gemma models, whose raw probabilities are overconfident. Temperature scaling improves Capability by 6.7 points for Gemma~4~31B and 10.0 points for Gemma~4~12B. After fine-tuning, the fitted temperatures are much closer to one, and the improvements are smaller.

For the trained Qwen models, many fitted temperatures are below one, which makes their predictions more confident. Using these temperatures directly reduces Capability from 81.2 to 81.0 for Qwen3.6-27B and from 66.6 to 64.6 for Qwen3.5-4B. This illustrates that temperatures fitted on our separate calibration set do not always transfer well to \jb. We therefore set any fitted temperature below one to one, preventing calibration from sharpening predictions. This restriction does not affect the Gemma models and is also applied to the public systems.

\begin{table}[h]
  \caption{\textbf{Temperatures of the released Jeff~1.1 models}, one per question type, fitted on held-out task families.}
  \label{tab:temps11}
  \centering\small
  \begin{tabular}{@{}lccc@{}}
    \toprule
    Model & Choice & Yes/no & Score \\
    \midrule
    Jeff-1.1-Large & 1.49 & 1.57 & 1.11 \\
    Jeff-1.1-Qwen-27B & 1.42 & 1.23 & 1.01 \\
    Jeff-1.1-Medium & 1.42 & 1.35 & 1.06 \\
    Jeff-1.1-Qwen-4B & 1.29 & 1.23 & 1.17 \\
    \bottomrule
  \end{tabular}
\end{table}

\begin{table}[h]
  \caption{\textbf{Raw readout vs.\ fitted temperatures} (\jb\ Capability; trained = mean of three seeds of the Jeff~1.0 recipe; temperatures
  never below 1).}
  \label{tab:temps}
  \centering\small
  \begin{tabular}{@{}lcccc@{}}
    \toprule
    & \multicolumn{2}{c}{Zero-shot} & \multicolumn{2}{c}{Trained} \\
    \cmidrule(lr){2-3}\cmidrule(lr){4-5}
    Model & Raw & Temperatures & Raw & Temperatures \\
    \midrule
    Gemma 4 31B & 72.9 & 79.6 & 81.9 & 84.4 \\
    Qwen3.6-27B & 78.2 & 79.0 & 81.2 & 81.2 \\
    Gemma 4 12B & 65.4 & 75.4 & 74.0 & 75.0 \\
    Qwen3.5-4B & 59.0 & 63.6 & 66.6 & 66.6 \\
    \bottomrule
  \end{tabular}
\end{table}

\begin{table}[h]
  \caption{\textbf{Fitted temperatures} per question type, fitted by NLL on our validation set (3,884 questions).
  Trained: mean of the three seeds of the Jeff~1.0 recipe, $\pm$ 1.96 standard errors. Italic: below 1;
  temperatures below 1 are set to 1 when serving (we never sharpen).}
  \label{tab:tempvalues}
  \centering\small
  \begin{tabular}{@{}llccc@{}}
    \toprule
    Model & Version & Choice & Yes/no & Score \\
    \midrule
    Gemma 4 31B & zero-shot & 5.96 & 7.24 & 7.24 \\
     & trained & \pmci{1.47}{0.09} & \pmci{1.31}{0.07} & \pmci{1.25}{0.07} \\
    \midrule
    Qwen3.6-27B & zero-shot & 1.38 & 1.52 & 1.85 \\
     & trained & \pmci{1.01}{0.03} & \pmci{\textit{0.83}}{0.03} & \pmci{\textit{0.77}}{0.04} \\
    \midrule
    Gemma 4 12B & zero-shot & 3.84 & 4.67 & 5.68 \\
     & trained & \pmci{1.19}{0.07} & \pmci{1.06}{0.07} & \pmci{1.10}{0.04} \\
    \midrule
    Qwen3.5-4B & zero-shot & 1.38 & 1.31 & 1.38 \\
     & trained & \pmci{\textit{0.86}}{0.03} & \pmci{\textit{0.79}}{0.07} & \pmci{\textit{0.76}}{0.02} \\
    \bottomrule
  \end{tabular}
\end{table}

\paragraph{Public systems.}
We apply the same calibration procedure to all competing systems, fitting a separate temperature for each question type using their predictions on our held-out calibration set. As with our models, we restrict temperatures to be at least one. For each system, we report whichever achieves higher Capability on \jb: its original predictions or its temperature-scaled predictions (Table~\ref{tab:temps-public}). Temperature scaling substantially improves several systems, including deck31b (+3.2), SemIf (+3.7), Winnow-12B (+3.8), and Torchcast-decision-12B (+2.8). However, it reduces Jev's Capability from 80.1 to 78.8, so we report Jev without additional calibration. Six other systems have all fitted temperatures below one and are therefore unchanged by our restriction.

\begin{table}[h]
  \caption{\textbf{Public systems with and without fitted temperatures} (\jb\ Capability). Table~\ref{tab:main} uses
  the higher of the two. Fitted temperatures before the rule; gray: below 1, so set to 1.}
  \label{tab:temps-public}
  \centering\small
  \begin{tabular}{@{}lccc@{}}
    \toprule
    System & As served & With temperatures & Fitted temperatures (choice / yes-no / score) \\
    \midrule
    StartLux-27B Instant & 88.5 & 88.5 & \textcolor{gray}{0.79} / \textcolor{gray}{0.42} / \textcolor{gray}{0.75} \\
    pplx-decider-v1.3 & 85.0 & 85.1 & \textcolor{gray}{0.94} / 1.06 / 1.06 \\
    Eikos-27B & 83.1 & 83.1 & \textcolor{gray}{0.61} / \textcolor{gray}{0.69} / \textcolor{gray}{0.79} \\
    StartLux-4B Instant & 82.2 & 82.2 & \textcolor{gray}{0.77} / \textcolor{gray}{0.56} / \textcolor{gray}{0.73} \\
    Quyet-1.0-Large & 81.8 & 81.7 & \textcolor{gray}{0.94} / \textcolor{gray}{0.82} / 1.06 \\
    deck31b & 78.1 & 81.3 & 1.68 / 2.59 / 2.37 \\
    AutoJev-27B & 80.5 & 80.5 & \textcolor{gray}{0.82} / \textcolor{gray}{0.77} / \textcolor{gray}{0.94} \\
    Chon-1.0-12B (no thinking) & 80.1 & 80.2 & 1.58 / 1.63 / 2.11 \\
    Jev 1.13.0 (API) & 80.1 & 78.8 & 1.73 / \textcolor{gray}{0.69} / 2.37 \\
    Torchcast-decision-12B & 75.6 & 78.4 & 1.12 / 4.22 / 1.41 \\
    Winnow-12B & 72.0 & 75.8 & 1.50 / 1.58 / 1.94 \\
    Cygnet & 75.3 & 75.6 & 1.09 / 1.19 / 1.50 \\
    Clef & 74.1 & 74.4 & \textcolor{gray}{0.87} / 1.15 / 1.12 \\
    pplx-decider-flash-v1.3 & 73.7 & 74.1 & 1.03 / 1.12 / 1.03 \\
    Wald-4B & 74.0 & 74.0 & \textcolor{gray}{0.61} / \textcolor{gray}{0.79} / \textcolor{gray}{0.53} \\
    Torchcast-decision-4B & 73.5 & 73.5 & \textcolor{gray}{0.84} / \textcolor{gray}{0.97} / \textcolor{gray}{0.89} \\
    Plumb-4B & 73.1 & 73.1 & \textcolor{gray}{0.84} / \textcolor{gray}{0.87} / \textcolor{gray}{0.94} \\
    Open-Jev-27B & 71.3 & 71.3 & \textcolor{gray}{0.33} / \textcolor{gray}{0.79} / \textcolor{gray}{0.37} \\
    decider-4b-v2 & 70.8 & 70.8 & \textcolor{gray}{0.97} / \textcolor{gray}{0.97} / \textcolor{gray}{0.77} \\
    SemIf & 58.9 & 62.7 & 1.30 / 1.30 / 1.41 \\
    Quyet-1.0-Medium & 58.9 & 58.8 & \textcolor{gray}{0.97} / \textcolor{gray}{0.92} / \textcolor{gray}{0.87} \\
    \bottomrule
  \end{tabular}
\end{table}

\paragraph{Where the temperatures act.} Table~\ref{tab:temps-subsets} breaks the effect down for Gemma~4~31B.
Temperatures never change the most likely option, so choice and yes/no competence stay the same, and they raise
calibration for every question type, most for the zero-shot readout (choice: 58.6 to 84.0). On score questions they
also lower competence, because the scorer measures it with the expected level of the predicted distribution, which a
temperature above 1 pulls toward the middle (zero-shot: 64.5 to 54.1). Across subsets, the zero-shot readout gains most
on synthetic (+6.6) and real items (+6.7), and loses on the negation pairs (100 to 94.3) and on the 35 items
with exact answer probabilities. After training, the changes per subset are small, except on these 35 items.

\begin{table}[h]
  \caption{\textbf{Gemma 4 31B with and without fitted temperatures, by question type and subset} (\jb; raw / with
  temperatures; trained = mean of three seeds). Subsets: Capability of their items under the same scorer.}
  \label{tab:temps-subsets}
  \centering\small
  \begin{tabular}{@{}lrcc@{}}
    \toprule
    & Items & Zero-shot & Trained \\
    \midrule
    \multicolumn{4}{@{}l}{\emph{Competence}} \\
    Choice & 1,178 & 84.5 / 84.5 & 86.4 / 86.4 \\
    Yes/no & 586 & 77.0 / 77.0 & 78.4 / 78.4 \\
    Score & 588 & 64.5 / 54.1 & 66.2 / 64.5 \\
    \midrule
    \multicolumn{4}{@{}l}{\emph{Calibration}} \\
    Choice & 1,178 & 58.6 / 84.0 & 81.2 / 92.6 \\
    Yes/no & 586 & 79.7 / 90.5 & 89.6 / 91.9 \\
    Score & 588 & 72.9 / 87.6 & 89.9 / 92.4 \\
    \midrule
    \multicolumn{4}{@{}l}{\emph{Capability by label source}} \\
    Code & 599 & 71.8 / 73.9 & 81.4 / 81.3 \\
    LLM-written & 1,443 & 72.3 / 78.9 & 80.5 / 83.1 \\
    Real requests & 310 & 71.9 / 78.6 & 77.9 / 77.0 \\
    \midrule
    \multicolumn{4}{@{}l}{\emph{Capability by family}} \\
    LLM routing & 694 & 76.3 / 81.9 & 82.7 / 83.6 \\
    Multilingual & 422 & 70.6 / 77.4 & 80.0 / 82.5 \\
    Fairness pairs & 236 & 62.2 / 63.9 & 65.0 / 65.6 \\
    Use cases & 213 & 68.4 / 73.8 & 78.3 / 79.4 \\
    Counting traps & 162 & 55.6 / 62.5 & 62.7 / 64.6 \\
    Negation pairs & 160 & 100.0 / 94.3 & 99.9 / 99.6 \\
    Option order & 120 & 77.2 / 77.7 & 83.7 / 82.2 \\
    Perturbation pairs & 110 & 48.9 / 52.0 & 60.8 / 63.2 \\
    Composite rules & 106 & 73.2 / 83.4 & 83.5 / 83.0 \\
    Calibration (generated) & 94 & 72.9 / 69.6 & 84.7 / 82.1 \\
    Calibration & 35 & 76.3 / 52.5 & 84.4 / 78.1 \\
    \midrule
    All & 2,352 & 72.9 / 79.6 & 81.9 / 84.4 \\
    \bottomrule
  \end{tabular}
\end{table}

\section{Answer readout ablations}\label{app:readout}

\paragraph{Probability outside the options.} The letter readout renormalizes the next-token probabilities of the $K$
option letters (Eq.~\ref{eq:readout}) and ignores all other tokens. For every zero-shot backbone the ignored mass is
small: at most 0.7\% on average on every set (Table~\ref{tab:offmass}). The most likely token is an option letter for
almost every item. The exceptions are ``Neither'' for 9 items of \jb\ (Gemma~4~12B), and ``NONE'' or ``None'' for 7
questions of our validation set whose options do not fit the state (Gemma~4~31B).

\begin{table}[h]
  \caption{\textbf{Next-token probability outside the option letters}, zero-shot (full vocabulary softmax
  at the answer position, including Gemma's soft-capping).}
  \label{tab:offmass}
  \centering\small
  \begin{tabular}{@{}lccccc@{}}
    \toprule
    & \multicolumn{3}{c}{\jb} & \pub & Our validation set \\
    \cmidrule(lr){2-4}\cmidrule(lr){5-5}\cmidrule(lr){6-6}
    Backbone & Mean outside & 99th percentile & Top token not an option & Mean outside & Mean outside \\
    \midrule
    Gemma 4 31B & 0.01\% & 0.2\% & 0 & 0.19\% & 0.24\% \\
    Qwen3.6-27B & 0.37\% & 4.9\% & 0 & 0.13\% & 0.67\% \\
    Gemma 4 12B & 0.52\% & 6.1\% & 9 & $<$0.01\% & 0.28\% \\
    Qwen3.5-4B & 0.09\% & 1.5\% & 0 & 0.05\% & 0.05\% \\
    \bottomrule
  \end{tabular}
\end{table}

\paragraph{Backbones without instruction tuning.} We also read out the pretrained versions of three backbones
(google/gemma-4-31B, google/gemma-4-12B, Qwen/Qwen3.5-4B-Base; Table~\ref{tab:base}). Our prompt does not work for them:
it uses the chat and thinking markers of the instruction-tuned models, which the pretrained models never saw. The
pretrained Gemma~4~31B puts 99.9\% of its next-token probability on tokens that continue them (``scratch'',
``thought'', ``thinking'', ``system'') and scores 37.2. We therefore also use a plain completion prompt, as is standard
for pretrained models: the same system text and JSON message as plain text, followed by ``Answer:'', reading the tokens
`` A'', `` B'', and so on. With this prompt, instruction tuning adds 14.9 points for Gemma~4~31B, 15.9 for Gemma~4~12B
and 4.1 for Qwen3.5-4B. The plain prompt suits the instruction-tuned 12B and 4B models less well than our prompt
(61.2 against 75.4, and 61.5 against 63.6), so the gap between a pretrained model and our zero-shot readout is larger
still. Training the pretrained 31B model with our recipe (and our prompt) gives 70.3, below the instruction-tuned model
without training (79.6).

\begin{table}[h]
  \caption{\textbf{Backbones with and without instruction tuning}, zero-shot letter readout with fitted temperatures. Chat: our prompt (Appendix~\ref{sec:requests}). Plain: the same text without chat markers, followed by
  ``Answer:''. Option mass: mean next-token probability on the option letters, on \jb. + recipe: trained with the Jeff~1.0
  recipe (400 steps, seed 1; the decision head reads out only the option letters).}
  \label{tab:base}
  \centering\small
  \begin{tabular}{@{}lllccccc@{}}
    \toprule
    & & & & \multicolumn{3}{c}{\jb} & \pub \\
    \cmidrule(lr){5-7}\cmidrule(lr){8-8}
    Backbone & Version & Prompt & Option mass & $I$ & $C$ & Capability & Accuracy (\%) \\
    \midrule
    Gemma 4 31B & instruction-tuned & chat & 99.99\% & 71.9 & 87.4 & 79.6 & 91.8 \\
     & instruction-tuned & plain & 69.6\% & 67.8 & 90.0 & 78.9 & 89.6 \\
     & pretrained & chat & 0.13\% & 7.9 & 66.4 & 37.2 & 41.1 \\
     & pretrained & plain & 66.1\% & 42.1 & 85.9 & 64.0 & 76.6 \\
     & pretrained + recipe & chat & -- & 58.4 & 82.1 & 70.3 & 81.0 \\
    \midrule
    Gemma 4 12B & instruction-tuned & chat & 99.5\% & 64.0 & 86.8 & 75.4 & 86.1 \\
     & instruction-tuned & plain & 74.4\% & 43.4 & 79.1 & 61.2 & 72.3 \\
     & pretrained & plain & 66.1\% & 14.0 & 76.7 & 45.3 & 58.0 \\
    \midrule
    Qwen3.5-4B & instruction-tuned & chat & 99.9\% & 46.1 & 81.2 & 63.6 & 80.1 \\
     & instruction-tuned & plain & 30.1\% & 40.5 & 82.6 & 61.5 & 72.7 \\
     & pretrained & plain & 55.6\% & 38.6 & 76.3 & 57.5 & 71.0 \\
    \bottomrule
  \end{tabular}
\end{table}

\paragraph{Instruction tuning and confidence.} In Table~\ref{tab:base-temps}, we show that with the same plain prompt, the instruction-tuned Gemma models are overconfident, while the pretrained ones are not. The mean
confidence in the chosen answer exceeds accuracy on \jb\ by 15 and 23 points on the instruction-tuned models, and the fitted temperatures are 2.7 to 6.0. The confidence-accuracy gaps for pretrained Gemma models are
within 2 points and need temperatures between 0.6 and 1.25. Interestingly, Qwen3.5-4B shows the opposite pattern with this prompt.

\begin{table}[h]
  \caption{\textbf{Confidence of instruction-tuned and pretrained backbones} (zero-shot letter readout, plain prompt).
  Gap: mean confidence in the chosen answer minus accuracy on \jb (without temperatures). Temperatures per question type fitted on held-out data (before the rule that keeps them at least one).}
  \label{tab:base-temps}
  \centering\small
  \begin{tabular}{@{}llcccc@{}}
    \toprule
    Backbone & Version & Gap & Choice & Yes/no & Score \\
    \midrule
    Gemma 4 31B & instruction-tuned & $+$14.7 & 4.24 & 5.68 & 5.96 \\
     & pretrained & $-$1.6 & 0.81 & 0.60 & 0.81 \\
    \midrule
    Gemma 4 12B & instruction-tuned & $+$23.4 & 2.73 & 3.32 & 3.84 \\
     & pretrained & $+$1.3 & 0.63 & 1.25 & 0.98 \\
    \midrule
    Qwen3.5-4B & instruction-tuned & $+$0.8 & 0.85 & 0.66 & 0.77 \\
     & pretrained & $+$10.2 & 1.13 & 1.13 & 0.98 \\
    \bottomrule
  \end{tabular}
\end{table}

\section{Latency}\label{app:latency}
\paragraph{Setup.} Every system is served on one NVIDIA H200 GPU with its authors' serving code and recommended
settings (StartLux AutoThink with gate 0.9, as in its published results), and called through JevBench's harness with
one question per request and one request at a time. Latency is the client-side time per decision on 500 \jb\ items (a
random sample, seed 0; the first 10 requests are left out as warm-up). Capability is measured on all 2,352 \jb\ items
with our metric, and in the column Cap.\ (JB) of Table~\ref{tab:latency} also with JevBench's (yes/no answers with $0.2 < P(\text{yes}) < 0.8$
count as wrong). Jev is called through its API, so its latency includes the network round trip.

\begin{figure*}[h]
  \centering
  \includegraphics[width=0.9\textwidth]{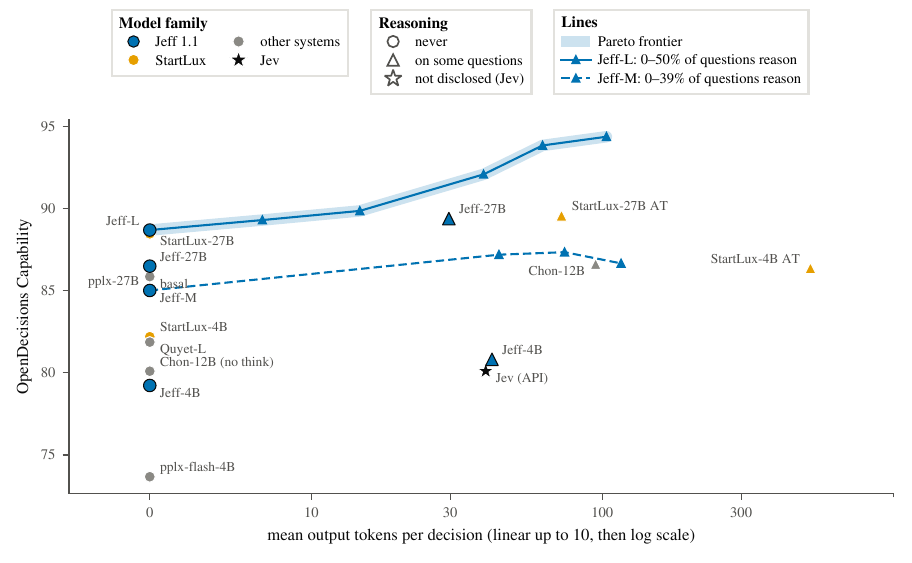}
  \caption{\textbf{\jb Capability against mean output tokens per decision}. The
  axis is linear up to 10 tokens and logarithmic above, so that the one-pass systems (0 tokens) are on the plot. Lines:
  Jeff-1.1-Large and Jeff-1.1-Medium with a growing share of questions that reason. Jev displayed with the length of its API response (no reasoning).}
  \label{fig:lat-tokens}
\end{figure*}

Our \href{https://github.com/j-geuter/Jeff}{serving code} for our models has the following options for inference optimization: \texttt{--compile} (the decision forward pass compiled
with \texttt{torch.compile}), \texttt{--reasoning-engine vllm} (the base model that writes the reasoning runs in vLLM on
the same GPU; the decision passes stay in jeff-serve) and, for the Qwen models, the flash-linear-attention kernels.
Table~\ref{tab:lat-speed} compares them with the default serving and with \texttt{--fast-reasoning} (the base model as a
second copy of the weights in transformers, with a static, compiled KV cache): the one-pass latency drops by 30--40\%
and the time per reasoning token by a factor of 3.5 (31B) to 9 (4B), at the same Capability within 0.5 points. In Figures~\ref{fig:lat-mean} and~\ref{fig:lat-tokens}, triangles mark systems that reason on some questions, and the lines show Jeff-1.1-Large and Jeff-1.1-Medium with a growing share of questions that reason, up to around 50\% and 39\%, respectively.

\begin{figure}[h]
  \centering
  \includegraphics{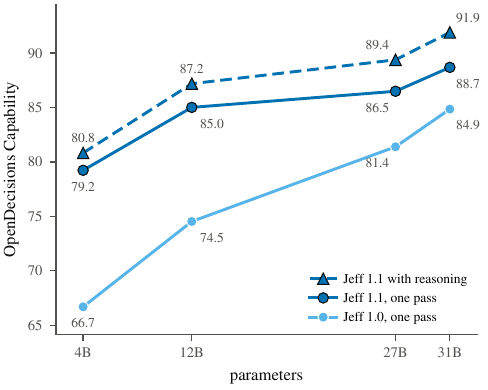}
  \caption{\textbf{Jeff 1.0 and Jeff 1.1 against model size, with Jeff 1.1 reasoning when unsure} (\jb, our metric).
  The reasoning line costs about the same for every model: 33 to 42 output tokens per question on average (31B 35, 27B
  33, 12B 38, 4B 42), with 10 to 20\% of the questions reasoning (15.5, 9.7, 11.2 and 20.0\%). The 31B, 27B and 4B use
  their released settings with the JevBench limits. The 12B's released setting costs about 100 tokens per question, so it
  uses its released shares scaled to a quarter; that run was chosen by its token count, not by its score.}
  \label{fig:lat-size-jeff-reason}
\end{figure}

\begin{table*}[h]
  \caption{\textbf{Default vs fast serving} of Jeff 1.1 (one H200): median latency per decision (ms) and \jb\ Capability
  in one pass (std: default; fast: \texttt{--compile}), and time per generated reasoning token (ms) and Capability with
  reasoning under the JevBench limits, the reasoning written by transformers (std: default; fast:
  \texttt{--fast-reasoning}) or by vLLM (\texttt{--reasoning-engine vllm}).}
  \label{tab:lat-speed}
  \centering\footnotesize
  \begin{tabular}{@{}lcccccccccc@{}}
\toprule
& \multicolumn{4}{c}{One pass} & \multicolumn{6}{c}{With reasoning (JevBench limits)} \\
\cmidrule(lr){2-5}\cmidrule(lr){6-11}
& \multicolumn{2}{c}{p50 (ms)} & \multicolumn{2}{c}{Cap.} & \multicolumn{3}{c}{ms/token} & \multicolumn{3}{c}{Cap.} \\
\cmidrule(lr){2-3}\cmidrule(lr){4-5}\cmidrule(lr){6-8}\cmidrule(lr){9-11}
Model & std & fast & std & fast & std & fast & vLLM & std & fast & vLLM \\
\midrule
Jeff-1.1-Large & 61 & 44 & 89.0 & 88.7 & 69 & 31 & 20 & 92.3 & 92.2 & 91.9 \\
Jeff-1.1-Qwen-27B & 93 & 57 & 86.8 & 86.5 & 73 & 24 & 15 & 89.5 & 89.2 & 89.4 \\
Jeff-1.1-Medium & 35 & 21 & 85.0 & 85.0 & 54 & 17 & 9 & 87.2 & 87.0 & 86.7 \\
Jeff-1.1-Qwen-4B & 48 & 30 & 79.3 & 79.2 & 36 & 8 & 4 & 81.0 & 81.0 & 80.8 \\
\bottomrule
\end{tabular}

\end{table*}

\begin{table*}[h]
  \caption{\textbf{Latency and Capability} on \jb\ (Cap.: our metric; Cap.\ (JB): JevBench's metric; latency on one
  H200 per decision; out tok: mean output tokens per decision; Jeff 1.1 with \texttt{--compile} and the vLLM reasoning
  engine). ``+R'' marks reasoning when unsure, with the
JevBench limits. Jev displayed with the length of its API response (no reasoning).}
  \label{tab:latency}
  \centering\footnotesize
  \begin{tabular}{@{}lrcrrrrrr@{}}
\toprule
System & Params & Reas. & Cap. & Cap. (JB) & p50 (s) & mean (s) & p90 (s) & out tok \\
\midrule
Jeff-1.1-Large +R & 31B & \cmark & 91.9 & 91.9 & 0.051 & 0.74 & 2.56 & 35 \\
StartLux-27B AutoThink & 27B & \cmark & 89.6 & 86.7 & 0.066 & 1.22 & 3.45 & 72 \\
Jeff-1.1-Qwen-27B +R & 27B & \cmark & 89.4 & 89.4 & 0.057 & 0.53 & 0.13 & 30 \\
Jeff-1.1-Large & 31B &  & 88.7 & 88.7 & 0.044 & 0.06 & 0.11 & 0 \\
StartLux-27B Instant & 27B &  & 88.5 & 83.1 & 0.040 & 0.05 & 0.08 & 0 \\
Jeff-1.1-Medium +R & 12B & \cmark & 86.7 & 86.7 & 0.031 & 1.12 & 4.01 & 116 \\
Chon-1.0-12B & 12B & \cmark & 86.6 & 86.6 & 0.078 & 3.75 & 19.84 & 95 \\
Jeff-1.1-Qwen-27B & 27B &  & 86.5 & 86.5 & 0.057 & 0.06 & 0.08 & 0 \\
StartLux-4B AutoThink & 4B & \cmark & 86.4 & 83.6 & 1.219 & 2.01 & 5.27 & 519 \\
basal-omni-1.0-ultra & 31B &  & 85.9 & 83.3 & 0.070 & 0.09 & 0.17 & 0 \\
pplx-decider-v1.3 & 27B &  & 85.0 & 81.5 & 0.064 & 0.07 & 0.10 & 0 \\
Jeff-1.1-Medium & 12B &  & 85.0 & 85.0 & 0.021 & 0.03 & 0.04 & 0 \\
Jeff-1.0-Large & 31B &  & 84.9 & 84.9 & 0.060 & 0.07 & 0.12 & 0 \\
vLLM-SR d3 & 27B &  & 84.5 & 80.5 & 0.064 & 0.07 & 0.10 & 0 \\
Eikos-27B & 27B &  & 83.1 & 78.2 & 0.044 & 0.06 & 0.10 & 0 \\
StartLux-4B Instant & 4B &  & 82.2 & 75.6 & 0.024 & 0.03 & 0.03 & 0 \\
Quyet-1.0-Large & 31B &  & 81.9 & 77.9 & 0.135 & 0.13 & 0.23 & 0 \\
Jeff-1.0-Qwen-27B & 27B &  & 81.4 & 81.4 & 0.094 & 0.11 & 0.17 & 0 \\
Jeff-1.1-Qwen-4B +R & 4B & \cmark & 80.8 & 80.8 & 0.029 & 0.18 & 0.92 & 42 \\
Jev 1.13.0 (API) & ? &  & 80.1 & 73.8 & 0.189 & 0.20 & 0.23 & 40 \\
Chon-1.0-12B (no thinking) & 12B &  & 80.1 & 80.1 & 0.049 & 0.05 & 0.06 & 0 \\
Jeff-1.1-Qwen-4B & 4B &  & 79.2 & 79.2 & 0.030 & 0.03 & 0.03 & 0 \\
Jeff-1.0-Medium & 12B &  & 74.5 & 74.5 & 0.035 & 0.04 & 0.06 & 0 \\
Wald-4B v2 & 4B &  & 74.0 & 74.0 & 0.027 & 0.03 & 0.05 & 0 \\
pplx-decider-flash-v1.3 & 4B &  & 73.7 & 69.9 & 0.033 & 0.03 & 0.03 & 0 \\
Torchcast-decision-4B & 4B &  & 73.5 & 73.5 & 0.013 & 0.01 & 0.02 & 0 \\
Jeff-1.0-Qwen-4B & 4B &  & 66.7 & 66.7 & 0.049 & 0.05 & 0.07 & 0 \\
Quyet-1.0-Medium & 4B &  & 58.8 & 54.8 & 0.140 & 0.16 & 0.27 & 0 \\
\bottomrule
\end{tabular}

\end{table*}

\end{document}